\documentclass[letterpaper, 10 pt, journal, twoside]{IEEEtran} 

\usepackage[pagebackref=false,breaklinks,colorlinks=true]{hyperref}
\usepackage{graphics} 
\usepackage{epsfig} 
\usepackage{mathptmx} 
\usepackage{times}
\usepackage{amsmath} 
\usepackage{amssymb} 
\usepackage{graphicx}
\usepackage{booktabs}
\usepackage{multirow}
\usepackage{makecell}
\usepackage{romannum}
\usepackage{caption}
\usepackage{subcaption}
\usepackage[dvipsnames]{xcolor}
\usepackage{MnSymbol}
\usepackage{hyperref}

\hypersetup{
    colorlinks=true,
    linkcolor=red,
    filecolor=magenta,      
    urlcolor=magenta,
}

\title{
Uncertainty Guided Policy for Active Robotic 3D Reconstruction using Neural Radiance Fields
}

\author{Soomin Lee$^{*}$, Le Chen$^{*}$, Jiahao Wang, Alexander Liniger, Suryansh Kumar$^\dag$, Fisher Yu
\thanks{$^\ast$Authors contributed equally.}
\thanks{Soomin Lee is with Oracle Labs Z\"urich,  Switzerland.}%
\thanks{Le Chen, Jiahao Wang, Suryansh Kumar, and Fisher Yu are with ETH Z\"urich, Switzerland.}%
\thanks{Alex Liniger is with Huawei Z\"urich, Switzerland.}%
\thanks{Note: This work was completed with VIS Group, ETH Z\"urich.}%
\thanks{$^\dag$Corresponding Author: Suryansh Kumar, ETH (\text{k.sur46@gmail.com}).}%
}

\newcommand{\etal}{\textit{et al}.}

\pdfoutput=1

\begin{document}
\maketitle
\thispagestyle{empty}
\pagestyle{empty}

\begin{abstract}
In this paper, we tackle the problem of active robotic 3D reconstruction of an object. In particular, we study how a mobile robot with an arm-held camera can select a favorable number of views to recover an object's 3D shape efficiently. Contrary to the existing solution to this problem, we leverage the popular neural radiance fields-based object representation, which has recently shown impressive results for various computer vision tasks. However, it is not straightforward to directly reason about an object's explicit 3D geometric details using such a representation, making the next-best-view selection problem for dense 3D reconstruction challenging. This paper introduces a ray-based volumetric uncertainty estimator, which computes the entropy of the weight distribution of the color samples along each ray of the object's implicit neural representation. We show that it is possible to infer the uncertainty of the underlying 3D geometry given a novel view with the proposed estimator. We then present a next-best-view selection policy guided by the ray-based volumetric uncertainty in neural radiance fields-based representations. Encouraging experimental results on synthetic and real-world data suggest that the approach presented in this paper can enable a new research direction of using an implicit 3D object representation for the next-best-view problem in robot vision applications, distinguishing our approach from the existing approaches that rely on explicit 3D geometric modeling.
\end{abstract}


\section{Introduction}
\label{sec:intro}
\begin{figure}
    \centering
    \setlength{\abovecaptionskip}{0.1cm}
    \setlength{\belowcaptionskip}{-0.4cm}
    \includegraphics[width=0.478\textwidth,trim={0cm 0cm 1cm 0cm},clip]{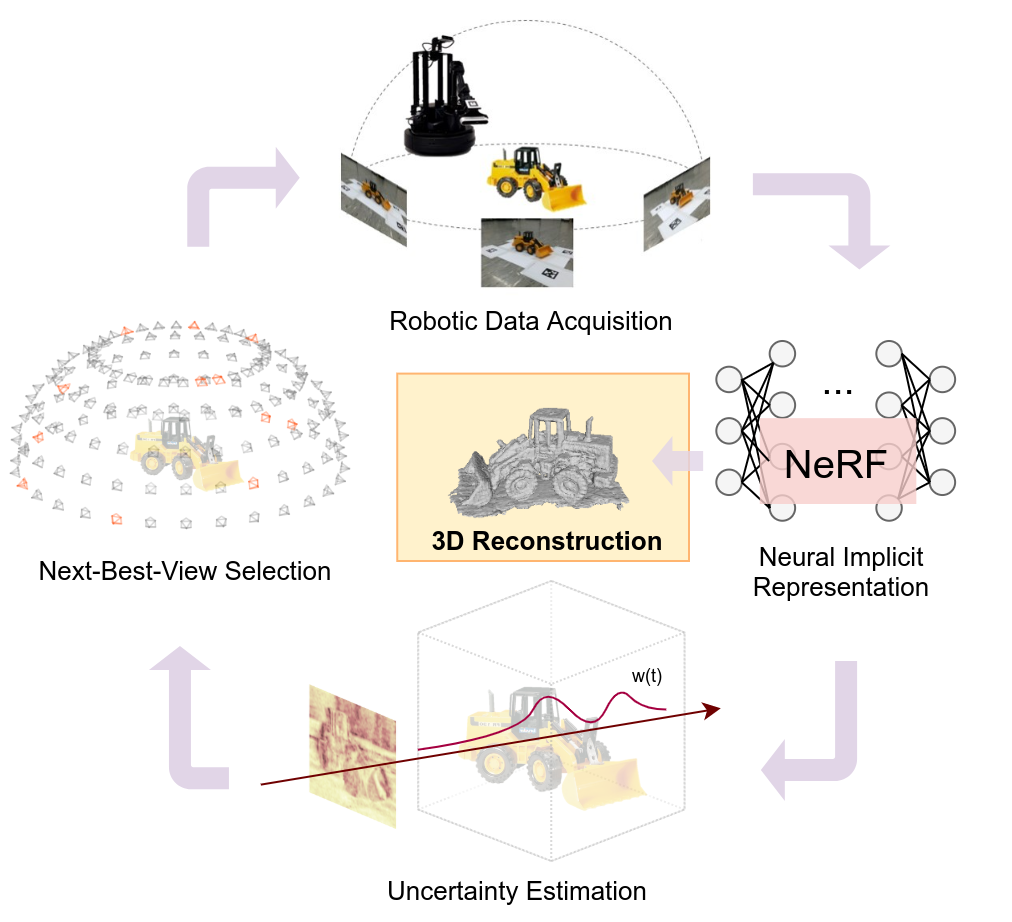}
    \caption[Our hypothesis]{\textbf{Overview.} We develop a robotic system that actively estimates the next-best-view for dense 3D reconstruction of an object leveraging the uncertainty modeling in implicit neural representation. For object representation, our work uses neural radiance fields \cite{mildenhall2020nerf} for its simplicity and notable performance on 3D shape representation.
    }
    \label{fig:hypothesis}
\end{figure}

Active vision-based robotic 3D reconstruction of an object using images or RGB-D sensors is a vital problem for robot vision and perception \cite{peralta2020next}\cite{scaramuzza}\cite{delmerico2018comparison}. The primary task of active vision in robotic systems is to skillfully operate the camera pose to capture as much information about the scene as possible. One practical approach to achieve this is by using a robot that can place its visual sensor such that the information gained for a given task is maximized \cite{Chen2011ActiveVI}\cite{zeng2020view}. That requires the robotic system to make planning decisions based on its state and current perceptual information of the environment without access to unseen information. This paper tackles the active robot vision problem popularly known as  next-best-view determination for object dense 3D reconstruction using multi-view images. A typical active robotic 3D reconstruction  {method} generally consists of three essential steps: \textbf{\emph{(i)}} Given the object's current information, proposals for the next possible view candidates are defined. \textbf{\emph{(ii)}} The best next view candidate is selected based on a specified criterion. \textbf{\emph{(iii)}} The robot maneuvers to the corresponding pose to obtain new 3D information about the object. These steps continue until no acceptable information gain is observed (see Fig.~\ref{fig:hypothesis}). Since  {the robot} has no access to the actual candidate views  {in step \textbf{\emph{(ii)}}}, it has to evaluate the unvisited view candidates, making the decision challenging and  {critical} in this pipeline.  {Thus}, the solution of this step is one of the main differentiating factors  {among} existing methods \cite{scaramuzza}\cite{devrim2017reinforcement}\cite{mendoza2020supervised}\cite{wang2019autonomous}\cite{wu2019plant}\cite{wu2014poisson}.

Available approaches that estimate the information gain for this problem work on explicit 3D representations such as pointcloud, voxel, etc., which are obtained via structure-from-motion (S\emph{f}M) or a calibrated RGB-D sensor.
However, it is well-studied that S\emph{f}M has limitations, and its suitability for achieving dense 3D reconstruction for robotic applications remains challenging \cite{bianco2018evaluating}\cite{schoenberger2016sfm}\cite{furukawa2009accurate}. Consequently, methods that rely on the fusion of depth maps coming from an active RGB-D sensor became popular \cite{curless1996volumetric}\cite{choi2015robust}\cite{newcombe2011kinectfusion}. Still, such methods are limited by the depth sensor acquisition range, depth sensor noise, and perceived depth accuracy, which is affected by the object's surface details and material type. Hence, both S\emph{f}M and RGB-D fusion methods have inherent drawbacks in dense 3D reconstruction, limiting their application in
active 3D reconstruction.

Recent advances in shape representation based on neural radiance fields, particularly NeRF \cite{mildenhall2020nerf}, have shown promising results  {for several computer vision tasks}. Using  {well-posed multi-view images}, NeRF can provide  {an} object's dense 3D reconstruction with favorable accuracy, overcoming the inherent limitations with S\emph{f}M \cite{schoenberger2016sfm}\cite{furukawa2009accurate}\cite{wu2011visualsfm} and depth fusion methods \cite{curless1996volumetric}\cite{newcombe2011kinectfusion}. 
Accordingly, we leverage the NeRF representation for the active robotic 3D reconstruction task. Contrary to the available methods that rely on explicit 3D representations, we explore the possibility of implicit neural shape representation to solve this problem. Nevertheless, due to the implicit nature of NeRF, the estimation of the information gain becomes even more challenging. We show that by computing the entropy of the weight distribution in the NeRF representation, we can reason about the information gain. We demonstrate with examples that such an approach is possible and can be effective.

\smallskip
\noindent
\textbf{Contributions.} To summarize, our key contributions are:

\begin{itemize}
    \item We introduce a new method of using implicit neural shape representation for the active robotic 3D reconstruction task.
    \item We show that entropy of the weight distribution of the color samples can be a suitable proxy for the uncertainty of the underlying 3D geometry, and we present an uncertainty guided policy using NeRF  {representation} for next-best-view  {selection}.
\end{itemize}

 {We provide extensive evaluation and comparison results on synthetic benchmark datasets to show the strength of our approach. Our experiments confirm the transferability of the proposed approach to real scenarios with superior results compared to the competing baselines.}

\section{Related Work}\label{sec:related}

\textbf{Active robotic 3D reconstruction.}
Available methods for solving this task generally rely on explicit 3D representations of the object. Isler~\etal \cite{scaramuzza} addresses information gain formulation in volumetric representations and compares the proposed metric with different methods in \cite{delmerico2018comparison}. Although the metric can be effective, it is a combination of several hand-crafted factors. Several other methods use point cloud for object representation, which is another widely used explicit 3D representation. Wu~\etal \cite{wu2014poisson} uses Poisson fields to get a confidence map of the current estimate to decide the part of an object that needs further scanning. The method focuses on the quality and accuracy of the recovered 3D surface, compromising on run-time instead. Wu~\etal replace the Poisson fields-based analysis with point completion network \cite{yuan2018pcn} to find incomplete parts of a scan, boosting up the speed but limiting their attention to plant phenotyping \cite{wu2019plant}.

%
In terms of robotic platforms that enable active 3D reconstruction, \cite{scaramuzza} has the most similar setting as ours. They use a wheeled mobile robot to move around an object and scan it with a camera.
\cite{wu2019plant} and \cite{wu2014poisson} both demonstrate their ideas on a robot but with a fixed base, using a scanner and an RGB-D sensor with multiple robot arms, respectively.


\smallskip
\noindent
\textbf{Neural 3D shape representations.} 
 {Neural implicit} shape representation via multi-layer perceptron (MLP) has recently gained popularity as an effective representation  {for 3D shapes} \cite{mildenhall2020nerf}\cite{icml2020_2086}\cite{Takikawa2021nglod}. Neural  {implicit} representations are independent of spatial resolution as geometry can be represented continuously without discretization and has a lower memory footprint. Earlier works for such a representation optimize a network to regress either the Signed Distance Function (SDF) or the occupancy function that takes 3D coordinates as an input \cite{mescheder2019occupancy}\cite{park2019deepsdf}. Although these methods can successfully represent 3D shapes, they require 3D supervision, restricting their  {use} to problems where the 3D geometry is unknown.

To our knowledge, this paper is an early attempt to build an active robotic 3D acquisition system based on a neural implicit representation of an object. While it is not yet common to adopt neural representations in robotics applications due to time constraints, a recent work \cite{SucarICCV2021imap} succeeded in using a neural representation to represent scenes in a real-time system. Furthermore, they demonstrated for the first time that a multi-layer perceptron could serve as the scene representation for an RGB-D SLAM system.


\smallskip
\noindent
\textbf{Volume and surface rendering for 3D reconstruction.}
In the past, S\emph{f}M, multi-view stereo, and depth-map fusion-based methods were widely used for active 3D acquisition tasks \cite{scaramuzza}\cite{schoenberger2016sfm}\cite{furukawa2009accurate}\cite{curless1996volumetric}\cite{schoenberger2016mvs}. However, as alluded to above, these classical approaches have limitations in the dense 3D reconstruction of an object.

Recently, neural volume and surface rendering methods have shown excellent 3D object reconstruction results.
For instance, DVR \cite{niemeyer2020differentiable} introduced a differentiable volumetric rendering formulation for multi-view 3D reconstruction using image data only. On the contrary, IDR \cite{yariv2020multiview} introduced a surface reconstruction approach leveraging a neural renderer that approximates the light reflected from the surface. Other recent approaches leverages implicit neural surfaces representation together with volume rendering idea for better 3D reconstruction \cite{oechsle2021unisurf}\cite{wang2021neus}\cite{yariv2021volume}. Nevertheless, among all, NeRF \cite{mildenhall2020nerf} turns out to be one of the most popular and widely used volume rendering approaches for object dense 3D acquisition and novel view-synthesis tasks.

NeRF is a simple yet effective volume rendering approach. It represents the continuous static scene as 5D neural radiance fields, parameterized by multi-layer perceptron (MLP). It demonstrated that regressing density and light fields via an MLP could yield photo-realistic rendering. Due to its remarkable ability to capture complex geometric details, it gathered significant interest from the community. NeRF led to several recent follow-up works that try to reduce the training time
\cite{yu2021pixelnerf}\cite{tensorf}, handling unknown or noisy camera poses \cite{wang2021nerfmm}\cite{lin2021barf}, adding depth supervision \cite{kangle2021dsnerf},or adding a notion of uncertainty
\cite{martin2021nerfw} {\cite{shen2021stochastic}}.  Since NeRF is simple, powerful, and forms the basis of all recent neural rendering works mentioned above, we choose NeRF methodology for this paper. Consequently, the idea presented in this paper can generalize to various other representations that stem from NeRF.

\section{Method}\label{sec:method}
Our work puts forward a policy formulation that selects the best candidate views for improving the 3D reconstruction in an active robot setting. Our approach infers the uncertainty from a proposed pose by synthesizing the novel view from NeRF-based shape representation. The rest of the section is organized as follows: Sec. \ref{ssec1:method}  {provides}  {an} overview of NeRF formulation \cite{mildenhall2020nerf}. Next, Sec. \ref{ssec2:method}  {introduces} our approach to model uncertainty. Lastly, Sec. \ref{subsec:policy},  {describes} our formulation for the uncertainty guided policy.

\vspace{-0.3cm}
\subsection{Preliminaries} \label{ssec1:method}
NeRF \cite{mildenhall2020nerf} models the continuous radiance fields of a static scene using a multilayer perceptron (MLP). It takes a set of images and encodes the scene as a volume density ($\sigma$) and color $\mathbf{c} = (r, g, b)$. NeRF renders each pixel of an image in a following way: Given a 3D point $(x, y, z)$ in the scene space and the ray direction parameterized by $(\theta, \phi)$ emitted from the camera's center of projection $\mathbf{o}$, NeRF learns an implicit function that approximates the scene $\sigma$ and $\mathbf{c}$ via an MLP$\big((x, y, z, \theta, \phi); \Theta\big) = (\sigma, \mathbf{c})$, where $\Theta$ is the parameter of an MLP network. Using $\sigma$ and $\mathbf{c}$ per scene point, we can render images from novel views via volume rendering \cite{kajiya1984ray}.

Consider a ray $r$ emanating from a camera position $\mathbf{o} \in \mathbb{R}^3$ in direction $\mathbf{d} \in \mathbb{R}^3, \text{where}~||\mathbf{d}||=1$. Volume rendering approximates light radiance by integrating the radiance along the ray $\mathbf{r}(t) = \mathbf{o} + t\mathbf{d}, t \geq 0$. Specifically, the expected color $\mathbf{C}(\mathbf{r})$ is computed using
\begin{equation}
    \mathbf{C}(\mathbf{r}) = \int_{0}^{\infty} T(t) \ \sigma(\mathbf{r}(t)) \ \mathbf{c}(\mathbf{r}(t),\mathbf{d}) \ dt ,
\label{eq:rendering}
\end{equation}
where $T(t)$ indicates the accumulated transmittance along the ray $r$ and is defined as
\begin{equation}
    T(t) = \exp{(-\int_{0}^{t} \sigma(\mathbf{r}(s)) \ ds)}.
\label{eq:transparency}
\end{equation}
It can be interpreted as the probability that a light particle traverses the segment [$\mathbf{o}, \mathbf{r}(t)$] without being bounced off. It's complement probability, denoted as the opacity $O$, is defined as $O(t) = 1 - T(t)$. The opacity $O$ is a monotonic increasing function with $O(0) = 0, O(\infty) = 1$. Thus, the opacity function $O$ can be regarded as a cumulative distribution function, and its derivative as a probability density function (PDF) \cite{yariv2021volume}
\begin{equation}
    w(t) = \frac{dO}{dt}(t) = T(t) \ \sigma(\mathbf{r}(t))
\label{eq:weight}
\end{equation}
The integrals in Eq.\eqref{eq:rendering} and Eq.\eqref{eq:transparency} can be estimated numerically using quadrature approximation \cite{max1995} as,
\begin{equation}
    \hat{\mathbf{C}}(\mathbf{r}) = \sum_{i=1}^{N} T_i \ (1-\exp{(-\sigma_i \delta_i)}) \ \mathbf{c}_i
\label{eq:quadrature}
\end{equation}
where, $T_i = \exp{(-\sum_{j=1}^{i-1} \sigma_j \delta_j)}$ and $\delta_i = t_{i+1} - t_i$ is the distance between adjacent samples. $\hat{\mathbf{C}}(\mathbf{r})$ in Eq.\eqref{eq:quadrature} can be viewed as a weighted sum of color samples $c_i$, and can be written as
\begin{equation}
    \hat{\mathbf{C}}(\mathbf{r}) = \sum_{i=1}^{N} w_i \mathbf{c}_i , \ \text{where}, \ w_i = T_i(1-\exp{(-\sigma_i \delta_i)})
\label{eq:weightedsum}
\end{equation}
The weight term $w_i$ in Eq.\eqref{eq:weightedsum} is a discrete approximation of the continuous weight expression in Eq.\eqref{eq:weight}, and can be derived using the approximation $\sigma_i \delta_i \approx (1-\exp{(-\sigma_i \delta_i)})$. Furthermore, if we define $p_i = \exp{(-\sigma_i \delta_i)}$, the discretized weight can be expressed as follows:
\begin{equation}
    w_i = (1-p_i) \prod_{j=1}^{i-1} p_j
\label{eq:interpret}
\end{equation}

\subsection{Ray-Based Volumetric Uncertainty}\label{ssec2:method}
As mentioned earlier, the neural radiance field representation has several advantages over other shape representations. At the same time, it can provide dense 3D reconstruction using multiview images only.  {Yet}, due to the implicit nature of the representation, it is not straightforward to directly operate on the explicit 3D shape and evaluate the 3D shape that the current network will yield. Moreover, the reasoning about the correctness of the object's volume density pivot around the multiview RGB color rendering values, making the inference about the 3D shape rather challenging.

We address  {such a} challenge  {in} potential next view  {selection} by analyzing the distribution of weight, $w(t)$ in Eq.\eqref{eq:weight}, along the rays of each pixel. Assuming we are looking for a solid surface, an ideal model should have a concentrated weight around the surface and nowhere else. This is also theoretically motivated since the weight term can be regarded as the derivative of the opacity, as shown in Eq.\eqref{eq:weight}. Thus, the weight distribution will have one clear peak if the network correctly learns about the surface. \cite{yariv2021volume} showed that the weight distribution indeed gets closer to a step function at the surface during training. We use the same observation to determine regions where NeRF has not yet learned a sufficiently good 3D representation. We argue that the regions with non-concentrated weights are where the 3D geometry can be improved.

\begin{figure}
\centering
\setlength{\abovecaptionskip}{-0.0cm}
\setlength{\belowcaptionskip}{-0.65cm}
\includegraphics[width=0.475\textwidth]{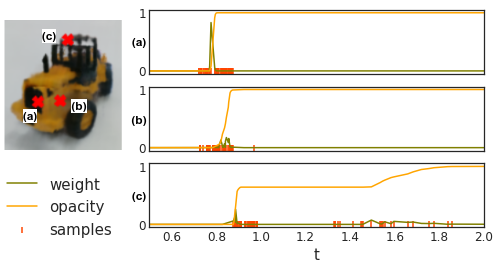}
\caption[Visualization of weight distributions]{\textbf{Key observation for the proposed ray-based volumetric uncertainty.} The weight $w(t)$, opacity $O(t)$, and sampled positions along rays are visualized. We can infer the uncertainty of the underlying reconstruction of the object by analyzing the weight distribution. (a) Accurate part: concentrated weight distribution with a clear peak. (b) Noisy part: concentrated weight distribution with a noisy peak. (c) Incomplete part: spread out distribution with multiple peaks.}
\label{fig:wdist}
\end{figure}

To confirm our hypothesis, we  {study} distributions of weight $w(t)$ along rays  {as shown in} Fig.(\ref{fig:wdist}). The first distribution shows a ray that intersects with a relatively accurate part, and has a clear peak. The second distribution shows a ray that intersects with a noisy part,  {which} has a noisy peak, but it is still concentrated. Finally, the third distribution shows a ray that intersects with an incomplete part, and the distribution has multiple peaks and is spread out. These results coincide well with our hypothesis. In summary, by examining how concentrated the weight distribution is, we can infer how certain the network is about  {the} ray.

Specifically, we quantify the degree of how concentrated a distribution is with entropy. Given a discrete random variable $X$, the entropy of $X$ is defined as:
\begin{equation}
    H(X) = -\sum_{i=1}^{n} P(x_i) \log P(x_i),
\label{eq:entropy}
\end{equation}
where $P(x_i)$ denotes the probability of the event $X=x_i$.
Entropy fits our purpose because evaluating whether a distribution has one sharp peak is in consonance with evaluating the uncertainty of a random variable the distribution yields. A uniform probability distribution yields the maximum entropy, while the entropy becomes zero when the outcome is always determined. Note that the weight $w(t)$ can be viewed as a PDF, as discussed earlier, so the definition of entropy can indeed be applied to the weight distributions.

\begin{figure}[t]
\setlength{\abovecaptionskip}{0.15cm}
\setlength{\belowcaptionskip}{-0.5cm}
\centering
\includegraphics[width=0.4\textwidth]{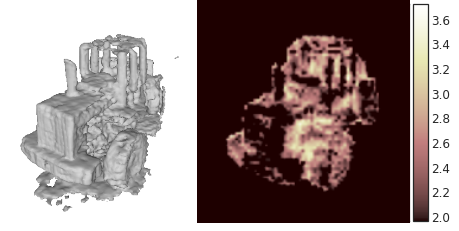}
\caption[Juxtaposition of 3D mesh and entropy map]{\textbf{Correlation between 3D mesh and entropy map.} Less precise parts in the 3D mesh (left) coincide with higher entropy parts in the object-masked entropy map (right). Brighter pixels indicate higher entropy values.}
\label{fig:3dentropy}
\end{figure}

Fig.(\ref{fig:3dentropy}) shows whether the uncertain regions align well with the inaccurate  {recovery} of a 3D mesh. We collect 60 images along a single horizontal circular trajectory around a toy loader using our robotic system and we train a model using those images. Then a 3D mesh is extracted from the model, which contains inaccurate regions due to insufficient coverage of the  {object} in the training data. One can confirm that the parts with high entropy match well with those not  {precisely reconstructed}, such as  {high-frequency regions}.

Our proposed uncertainty estimation has several advantages: First of all, the idea can be  {directly} generalized to different works that are based on neural rendering. Estimating the weight distribution along each ray is one of the processes that commonly exist in every work that leverages neural rendering. Next, it is simple and easily applicable because it does not require any additional training or changes in the network. Finally, it provides a metric that can be evaluated on the combined effects of different sources of uncertainty, such as deficiency of data or geometric complexity. Accordingly, we can avoid reasoning about different criteria we need to consider, eliminating the need to use heuristics as in \cite{scaramuzza}.

 {Several recent works proposed methods to identify the uncertainty in NeRF~\cite{martin2021nerfw}\cite{shen2021stochastic}. NeRF-W~\cite{martin2021nerfw} models static and transient elements separately in order to handle uncontrolled images, and the notion of uncertainty mainly serves as an attenuation factor for the transient elements rather than the uncertainty of 3D reconstructions of static scenes.
S-NeRF \cite{shen2021stochastic} learns to encode the posterior distribution over all the possible radiance fields modeling the scene and obtains the uncertainty estimates by sampling, following a Bayesian approach. However, both of them require models to be modified, while our method can be directly plugged into other works that are based on neural rendering.}

\smallskip
\noindent
\textit{Note:}
The proposed approach, however, is restricted to non-transparent objects with solid surfaces. Volume rendering-based models such as NeRF can model transparency. Yet, it is not straightforward to recover the 3D surface of the transparent glass. The extraction of 3D geometry usually relies on the Marching Cube \cite{mcubes} algorithm, which can easily remove low volume density regions together with noise.  {Hence, our work is suitable for non-transparent objects}.

\vspace{-0.1cm}
\subsection{Uncertainty Guided Policy}\label{subsec:policy}
For  efficient active robotic 3D reconstruction, the fundamental task is to decide which views to scan next. As highlighted in the introduction, the challenge comes from having no access to the view at the decision time, i.e., step \textbf{\emph{(ii)}}. Our proposed ray-based volumetric uncertainty estimation approach allows us to infer the importance of adding a novel view image via its uncertainty estimate to address such a challenge. Thus, by design, it is straightforward to convert the ray-based volumetric uncertainty estimator into a policy by considering the mean entropy of a candidate view as a proxy to the information gain this view can bring.

When we select images based on the proposed implicit volumetric uncertainty, we take the mean of the entropy values across all pixels to be the representative value for an image. We find the mean values sufficient for our task of view selection; however, one can potentially reason about local uncertainties using the information since we compute the uncertainty measure for each pixel.

\begin{figure}
\setlength{\abovecaptionskip}{-0.0cm}
\setlength{\belowcaptionskip}{-0.5cm}
    \centering
    \includegraphics[width=0.37\textwidth]{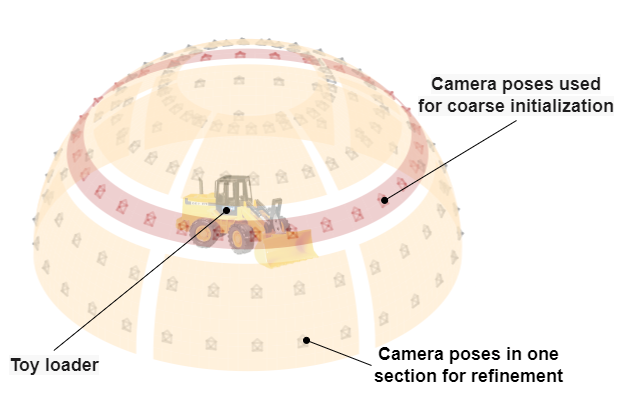}
    \caption[Division of the view space]{\textbf{Region clustering.} The view space defined on a hemisphere is divided into several sections to locally determine additional training samples. The middle part indicates the circular trajectory where the initial poses are sampled from, and thus is excluded when clustering camera poses for additional training.}
    \label{fig:section}
\end{figure}

In this work, we investigate a coarse-to-fine reconstruction approach, where we start with a coarse set of images and select views to improve the initial reconstruction. Consequently, we introduce region clustering, as shown in Fig.(\ref{fig:section}),  {which consists} of a region where the initial views are selected and several additional regions where further views can be selected.  {After a coarse reconstruction using the initial views, each iteration selects the view with the highest mean entropy of each sector, and the robot collects the corresponding view}. These views are then added to the dataset, based on which the model is refined.
Without region clustering, the overall acquisition process will be much longer to cover the object geometry from different viewpoints, as we will get only one view every iteration. Alternatively, selecting the top-k most uncertain viewpoints without updating the model may result in choosing a group of similar viewpoints. However, this is not optimal since only a subset of them may be sufficient to reduce the uncertainty in the region, and using similar views could lead to overfitting. On the other hand, splitting the view selection space into regions is a simple yet effective step that helps select diverse views, which can potentially be used to incorporate a prior based on domain knowledge. We discuss other selection policies in Sec. \ref{sec:ablation}.

\section{Experiment}
\label{sec:experiment}

We evaluate our uncertainty guided policy on both synthetic and real-world data covering several types of objects. For clarity, experimental setup and results on synthetic and real-world dataset are described separately in Sec. \ref{subsec:synthetic} and Sec. \ref{subsec:real_exp}, respectively.

\smallskip
\noindent
\textbf{Implementation Details.}
We define the view space to be a hemisphere surrounding the object and acquire images of the object from five circles with different radii on the hemisphere (see Fig.(\ref{fig:section})). Thirty candidate poses are defined for each of the five horizontal circles on the hemisphere, resulting in 150 camera poses. Then as an initialization, we train a NeRF model using images only from the middle circle. We use six images for initialization  {(coarse reconstruction)} in experiments with synthetic data, while we use 15 images for real-world experiments. We use this setup as a few images are enough to get a high-quality 3D mesh for synthetic data. Still, significantly more images are required for real objects, showing the importance of both synthetic and real-world experiments. Then we divide the hemisphere into 12 sections, as shown in Fig.(\ref{fig:section}) for region clustering. The hemisphere is divided into the upper and lower half with respect to the middle circle that contains the initial training images, and each half is further divided into six groups according to their azimuthal angles. Therefore, in total, we have 12 sections and one middle circle, and we select 12 additional images {, one} from each section in each iteration.

For training a NeRF model, we use the official code provided by the authors \cite{mildenhall2020nerf}. We use $64$ samples for the `coarse' network and $128$ samples for the `fine' network. When evaluating the entropy of the weight distributions, we downsample the images with a factor of $4$ to speed up the process. After selecting additional images, we initialize the network with the model from the initialization step and  {refine the model} further with the updated training set.

\smallskip
\noindent
\textbf{Evaluation Metric.} For 3D reconstruction evaluation, we use the popular F-score metric \cite{Knapitsch2017tank}. The F-score is the harmonic mean of precision and recall at a certain threshold $d$. The precision quantifies the accuracy of the reconstruction, and it can be maximized by producing a very sparse set of precisely localized landmarks for instance. The recall quantifies the completeness of the reconstruction, and it can be maximized by densely covering the space with points.

\begin{figure}
\setlength{\belowcaptionskip}{-0.56cm}
    \centering
    \includegraphics[width=0.36\textwidth]{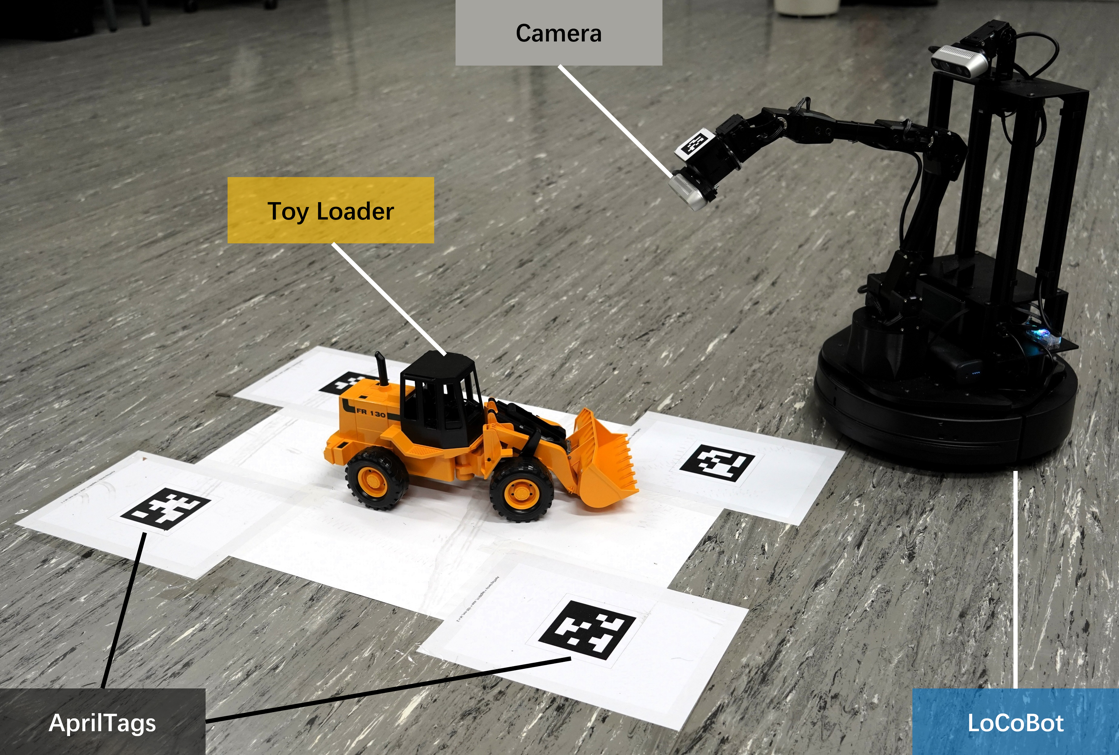}\caption[Experimental setup]{\textbf{Experimental setup for real-world object reconstruction.} The toy loader is placed in the middle of 4 AprilTags \cite{apriltag} for localization. The gripper of LoCoBot~\cite{gupta2018robot} is replaced with a camera.}
    \label{fig:fourtags}
\end{figure}

\smallskip
\noindent
\textbf{Hardware.}
We use LoCoBot, a low-cost mobile manipulator hardware platform to perform active robotic 3D reconstruction \cite{gupta2018robot}. It has a wheeled mobile base with two degrees of freedom (DoF), a manipulator with 5 DoF, and a camera attached to the top. Initially, the robot had a gripper at the end of the manipulator, but it was replaced with another camera to allow the robot to perform exact 3D reconstruction.  Note that we \emph{do not} use depth information from the RGB-D sensor. The adjusted hardware and the experimental setup are shown in Fig.(\ref{fig:fourtags}). Using the camera on top, we localize the robot with respect to AprilTags \cite{apriltag} to control the robot and to position the camera on the arm to   {acquire well-posed images}.

\vspace{-0.2cm}
\subsection{3D Reconstruction of Synthetic Objects}
\label{subsec:synthetic}

\begin{table*}[ht]
\centering
\caption{\textbf{F-score of synthetic object reconstruction.} With access to the ground truth meshes, we show a quantitative comparison against the baselines for the reconstructed geometry on 4 different synthetic objects. Our method performs the best among all the selection policies. We also report the results of the mesh reconstructed with all 150 images for reference. Bold numbers are only considering policies.  {The computation time for the next-best-view selection using each policy: {Random and Heuristic- less than a second, Similarity (GT)- 15 sec., VI\cite{scaramuzza}- 13.8 sec., Similarity and Ours- approx. 5 min}}.}
\footnotesize{
\setlength\tabcolsep{9pt}\
\begin{tabular}{lccccccccc}
    \toprule
    \multirow{2}{0.8cm}[-.6em]{Object}
    & \multirow{2}{*}[-.6em]{Initialization}
    & \multirow{2}{*}[-.6em]{Pure Random}
    & \multicolumn{6}{c}{Policy}
    & \multirow{2}{*}[-.6em]{All Images} \\
    \cmidrule(lr){4-9}
    & & & Random & Heuristic & Similarity & Similarity (GT) & VI \cite{scaramuzza} & \textbf{Ours} &  \\
    \midrule
        \textit{Lego} & 0.3549 & 0.3682 & 0.3909 & 0.3959 & 0.3873 & 0.3710 & 0.1824 & \textbf{0.4101} & \textit{0.4374} \\
        \textit{Chair} & 0.1285 & 0.1696 & 0.1831 & 0.1615 & 0.1772 & 0.1836 & 0.0959 & \textbf{0.1858} & \textit{0.2142} \\
        \textit{Drums} & 0.2778 & 0.2229 & 0.2766 & 0.2700 & 0.2732 & 0.2687 & 0.1548 & \textbf{0.2853} & \textit{0.3793 }\\
        \textit{Ficus} & 0.1788 & 0.2557 & 0.2622 & 0.2666 & 0.2676 & 0.2630 & 0.1735 & \textbf{0.2705} & \textit{0.3781} \\
    \bottomrule
\end{tabular}

}
\label{tab:synthetic_result}
\vspace{-0.2cm}
\end{table*}

\newcommand{\resultsfigwidth}{0.9in}
\newcommand{\resultscropwidth}{0.69in}

\newcommand{\cropship}[2][1]{
  \makecell{
  \includegraphics[trim={180px 100px 180px 90px}, clip, width=\resultscropwidth]{#2} \\
  \includegraphics[trim={180px 130px 180px 70px}, clip, width=\resultscropwidth]{#1}
  }
}

\newcommand{\cropficus}[2][1]{
  \makecell{
  \includegraphics[trim={175px 95px 175px 85px}, clip, width=\resultscropwidth]{#2} \\
  \includegraphics[trim={220px 240px 310px 80px}, clip, width=\resultscropwidth]{#1}
  }
}

\begin{figure*}[t]
\setlength{\abovecaptionskip}{0.1cm}
\setlength{\belowcaptionskip}{-0.4cm}
\centering
\scriptsize
\begin{tabular}{@{}c@{}c@{}c@{}c@{}c@{}c@{}c@{}c@{}c@{}}
\makecell[c]{
\includegraphics[trim={180px 100px 180px 0px}, clip, width=\resultsfigwidth]{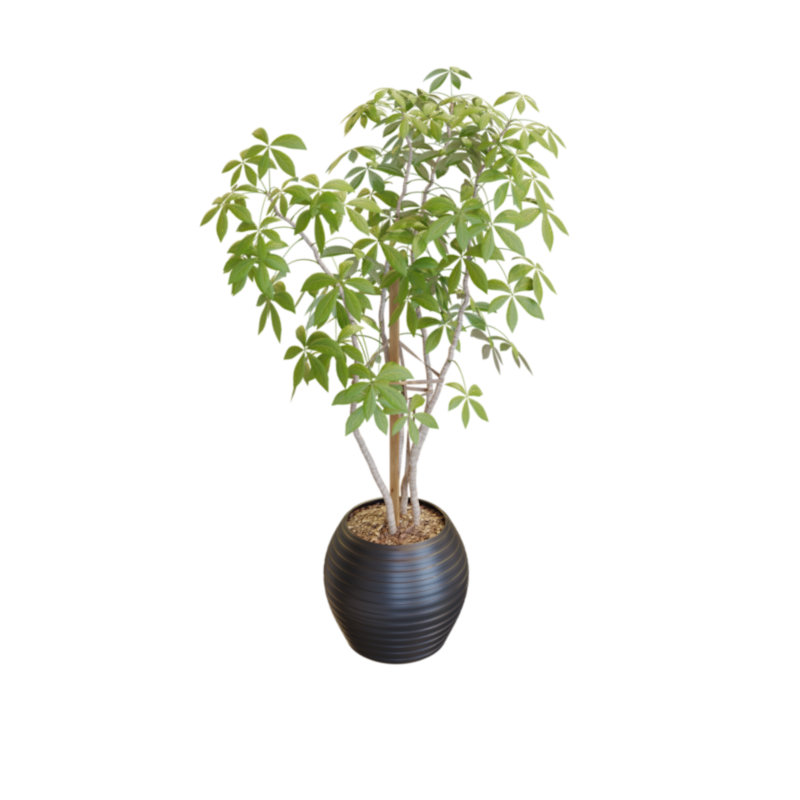}
\\
Ficus
}
&
\cropship[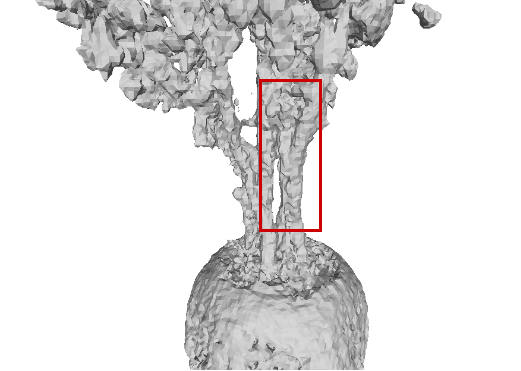]{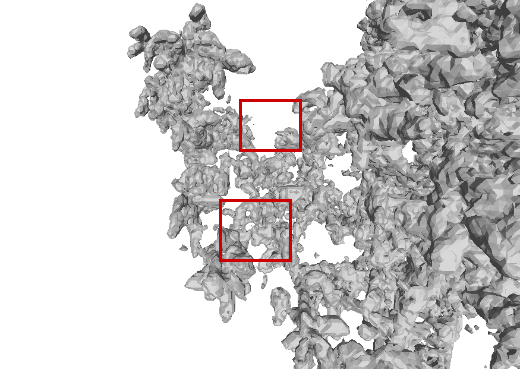} &
\cropship[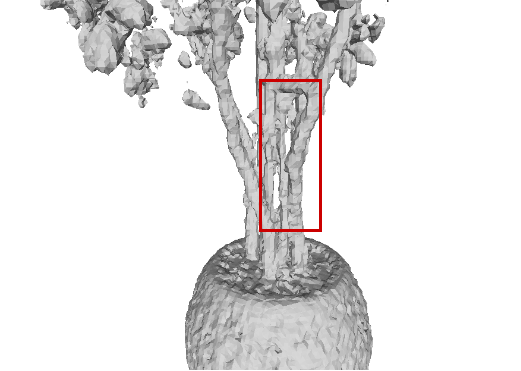]{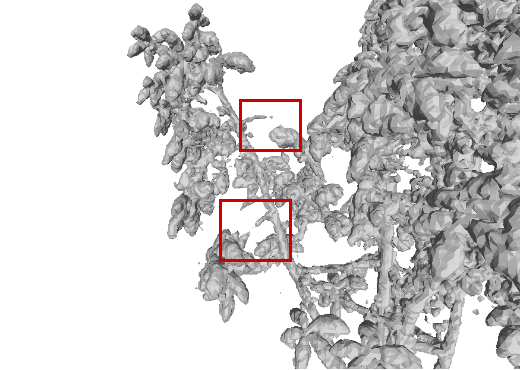} &
\cropship[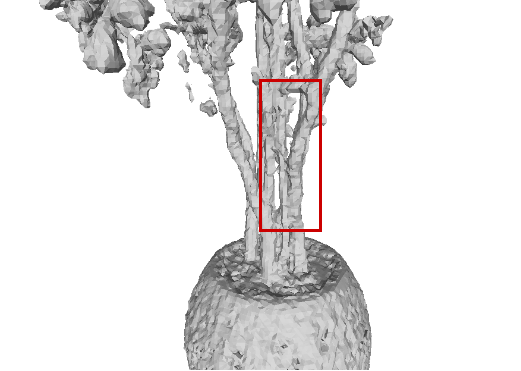]{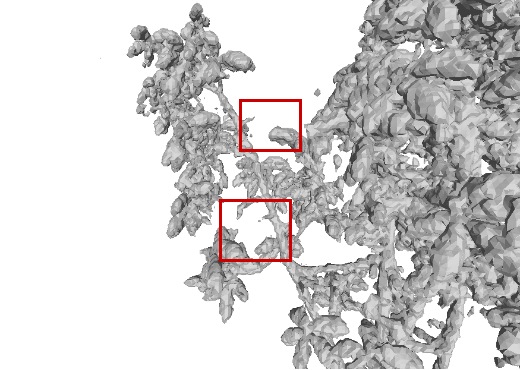} &
\cropship[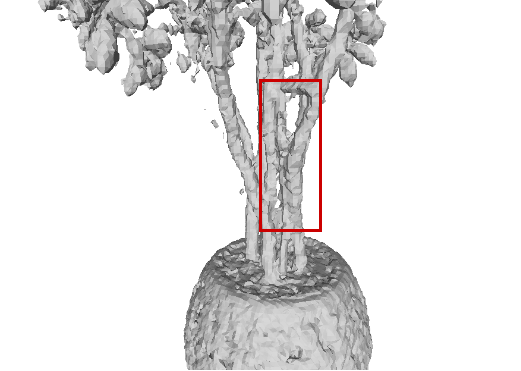]{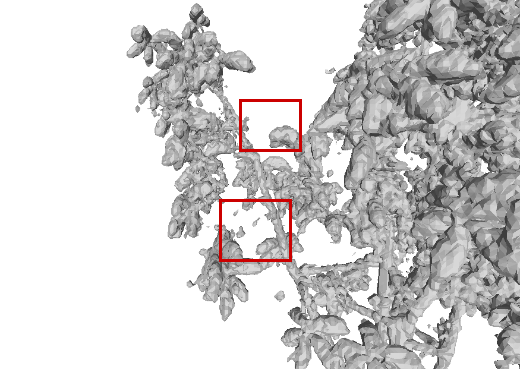} &
\cropship[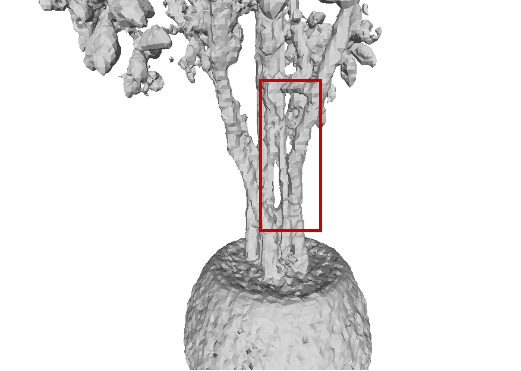]{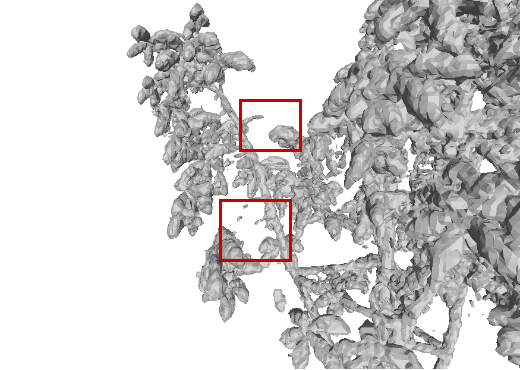} &
\cropficus[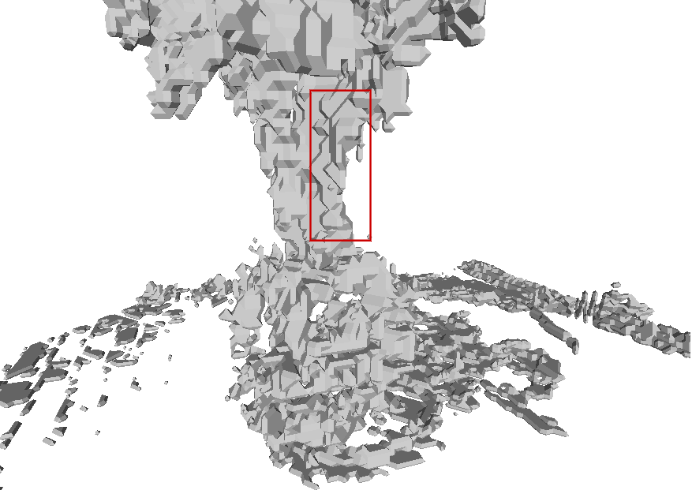]{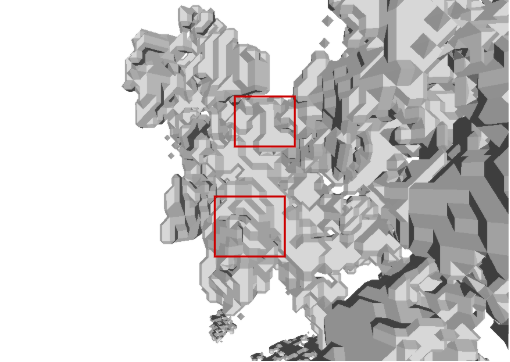} &
\cropship[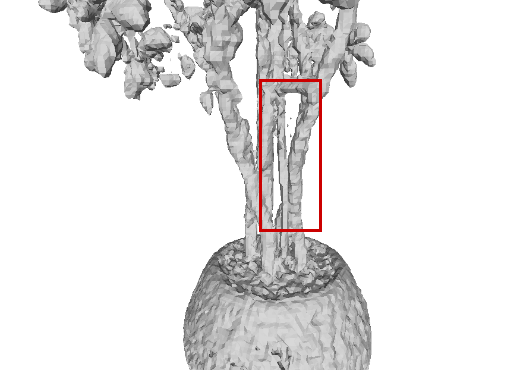]{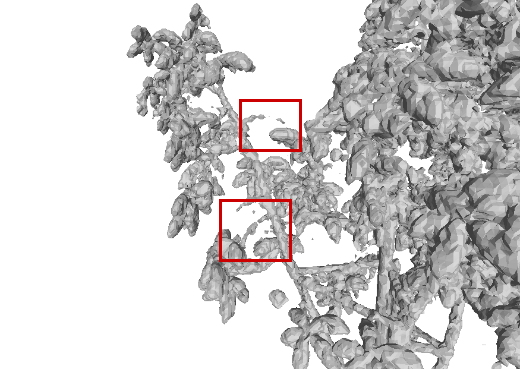} &
\cropship[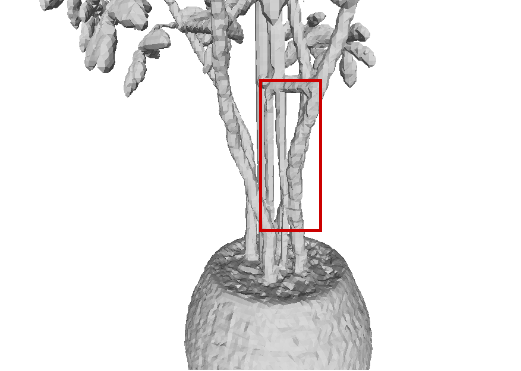]{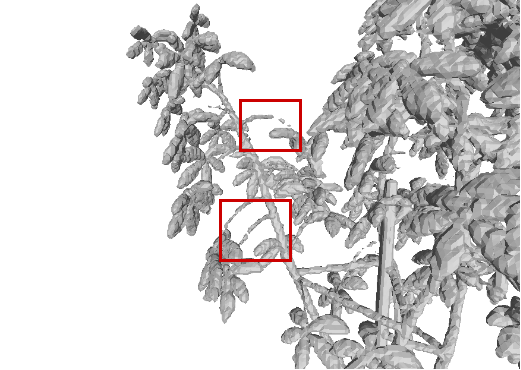} \\
& Initialization & Random & Heuristic & Similarity & Similarity (GT) & VI~\cite{scaramuzza} & Ours & \text{All Images}
\end{tabular}
\caption{\textbf{Comparisons on 3D meshes for \textit{Ficus}.} We show a qualitative comparison against the baselines for the reconstructed geometry on \textit{Ficus}, which has high frequency details. Our method captures the fine geometry well compared to the other policies. Note that the VI~\cite{scaramuzza} method yields voxel representations.}
\label{fig:ficus}
\end{figure*}

\noindent
\textbf{(a) Datasets.}
For this experiment, we use the NeRF Blender dataset \cite{mildenhall2020nerf}.
We select four objects, namely \textit{Lego}, \textit{Chair}, \textit{Drums}, and \textit{Ficus} to generate dataset for our robotic setup. Note that the transparent surfaces of the \textit{Drums} model are removed to satisfy our assumption. We render 150 images according to our experimental setup mentioned before.

\noindent
\textbf{(b) Baselines.} For each object, we report the results of the initial model trained with 6 images from the middle circle and the model trained with all 150 images from the hemisphere. We also present 5 different next-best-view selection policies as baselines. \textit{\textbf{(i)} Random policy}: a pose is selected randomly within each section. \textit{\textbf{(ii)} Heuristic policy}: the middle pose of each of the 12 sections is selected.
\textit{\textbf{(iii)} Similarity policy}: within each section, a pose with the lowest image similarity to the initial training data is selected.
Using the initial NeRF model, we synthesize candidate views and compute the cosine similarity between the feature vector of the synthesized views and the initial training images. The feature vector is obtained with ResNet-18~\cite{he2016deep} pre-trained on ImageNet~\cite{deng2009imagenet}.
\textit{\textbf{(iv)} Similarity (GT) policy}: a pose is selected in the same way as \textit{\textbf{(iii)}}, but the ground truth images are used for feature extraction instead of the synthesized views. While the baselines \textit{\textbf{(i)}}-\textit{\textbf{(iv)}} all are based on NeRF, we additionally compare our approach with a volumetric active 3D reconstruction method, denoted as \textit{\textbf{(v)} Volumetric Information (VI) policy} \cite{scaramuzza}.  {Note that this method uses stereo images as input.} \textit{\textbf{(vi)}  Pure Random}:  {together with the aforementioned policies, we also present} a pure random baseline where images are randomly chosen over the entire view space rather than within each section.

\noindent
\textbf{(c) Results.} We run each baseline view selection policy based on the initial model trained with 6 images to select one image within each section. It means that after one iteration, we have 12 additional images to refine the reconstruction. We report the reconstruction results after one iteration of different view selection policies in Table~\ref{tab:synthetic_result}. From Table~\ref{tab:synthetic_result}, we can see that for all the synthetic objects we have tested on, our uncertainty guided policy obtains the highest F-score against all the baselines. Note that all NeRF-based baseline policies, except for the pure random baseline (i.e., \textbf{\textit{(vi)}}), are relatively similar since they all get 18 views as an input which are reasonably well distributed. Further, our method achieves improvement up to 30\% compared to the pure random baseline. On the whole, these statistical results demonstrate that our policy selects the best view from each region on the hemisphere.

Additionally, the visual similarity policy study shows that the ray rendering uncertainty is more informative than the visual features. Moreover, VI~\cite{scaramuzza} baseline by far achieves the lowest quality reconstruction.  This affirms the suitability of our choice of using implicit neural implicit volumetric representation and modern continuous view-synthesis approaches like NeRF to solve this problem. Fig.(\ref{fig:ficus}) shows the qualitative results compared to the defined baselines. Clearly, the reconstructions from our method better capture the fine geometric details and coherent overall global shape.

\subsection{3D Reconstruction of Real-World Objects}
\label{subsec:real_exp}

\begin{figure*}[!hbt]
\setlength{\abovecaptionskip}{0.72cm}
\setlength{\belowcaptionskip}{-0.7cm}
    \centering
     \begin{subfigure}[b]{0.24\textwidth}
         \centering
         \includegraphics[width=\textwidth]{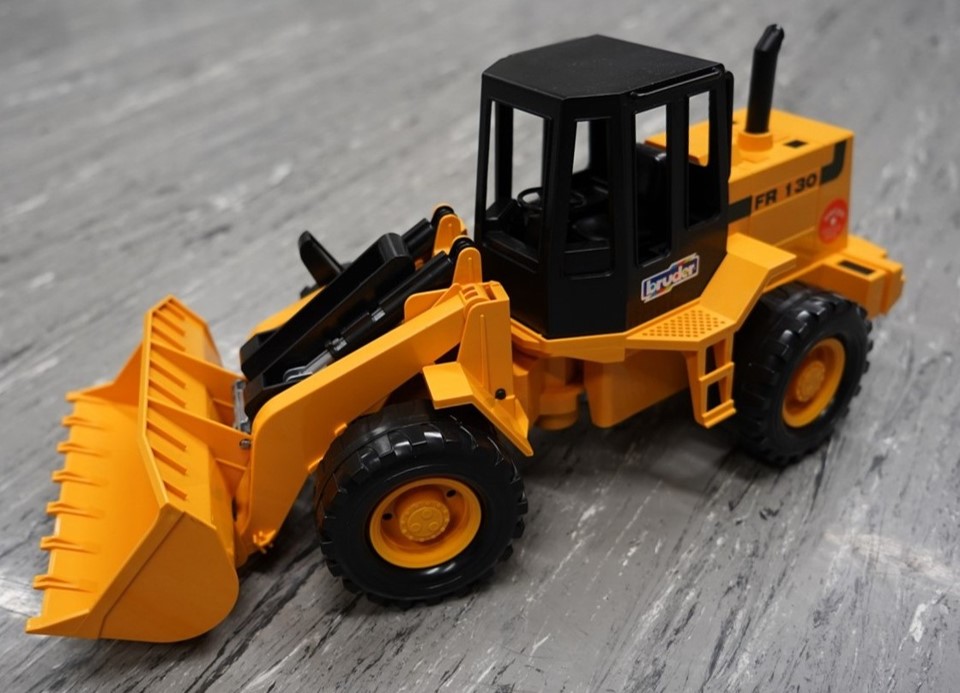}
         \caption{Ground Truth}
         \label{fig:real_toy_img}
     \end{subfigure}
     \begin{subfigure}[b]{0.24\textwidth}
         \centering
         \includegraphics[width=\textwidth]{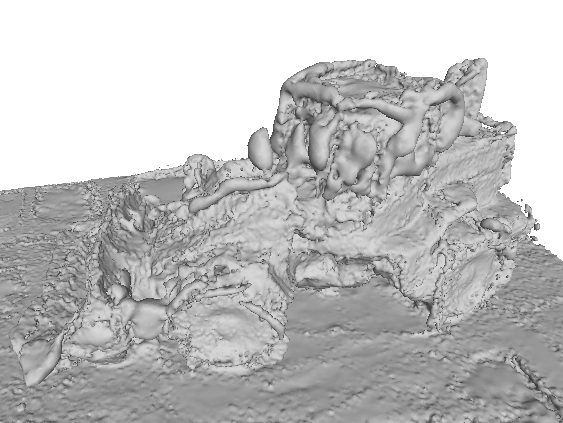}
         \caption{COLMAP \cite{schoenberger2016sfm}}
         \label{fig:colmap}
     \end{subfigure}
    \begin{subfigure}[b]{0.24\textwidth}
         \centering
         \includegraphics[width=\textwidth]{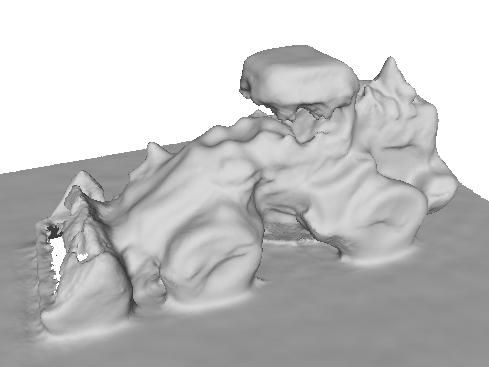} 
         \caption{Open3D (TSDF-Fusion \cite{curless1996volumetric})}
         \label{fig:tsdf}
     \end{subfigure}
    \begin{subfigure}[b]{0.24\textwidth}
         \centering
         \includegraphics[width=\textwidth]{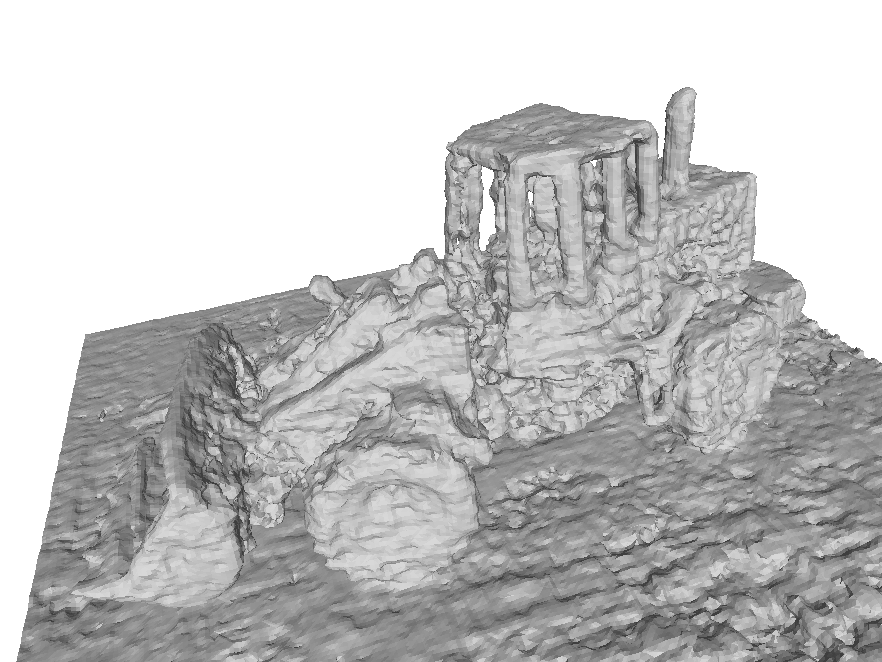} 
         \caption{Ours}
         \label{fig:ours}
     \end{subfigure}
    \caption[Experimental setup]{\textbf{Reconstruction results of a toy loader using classical methods and our approach with real-world data.} The toy loader is reconstructed using COLMAP \cite{schoenberger2016sfm, schoenberger2016mvs} with 150 images, the reconstruction system provided by Open3D \cite{choi2015robust, Park2017ColoredPC} with 2000 images, and our approach with 27 images (15 for initialization, 12 for refinement).}
    \label{fig:realresult}
\end{figure*}

\noindent
\textbf{(a) Datasets.} We use our robotic system to acquire images of real-world objects. For this experiment, we used the toy loader shown in Fig.\ref{fig:real_toy_img} as the target object and acquired images at $640 \times 480$ resolution. We compute the camera poses using COLMAP \cite{schoenberger2016sfm}\cite{schoenberger2016mvs}. 
Similar to the synthetic data experiment, we define 150 candidate camera poses. The robot takes about 1.2 minutes to collect 15 images from the middle circle to initialize the coarse 3D shape representation. It takes another 1 minute to collect 12 more images for refinement using the proposed uncertainty guided policy.

\noindent
\textbf{(b) Baselines.} We reconstruct the toy loader under two baseline settings: (\textbf{\Romannum{1}}) COLMAP and screened Poisson surface reconstruction \cite{poisson}, (\textbf{\Romannum{2}}) a reconstruction algorithm provided by Open3D \cite{choi2015robust}\cite{Park2017ColoredPC} that uses RGB-D image sequences. For baseline (\textbf{\Romannum{1}}), we use all 150 images for reconstruction. When we use the same 27 images that are used for our method, the structure-from-motion pipeline fails because the images contain large rotational changes in their views. For baseline (\textbf{\Romannum{2}}), an RGB-D image sequence is required and hence an RGB-D image sequence consisting of 2000 images is used.

\noindent
\textbf{(c) Results.} The resulting 3D meshes are shown in Fig.(\ref{fig:realresult}). Conceivably, 3D geometry recovered using our method has the highest quality with fewer images. It can be observed the COLMAP fails to reconstruct the fine surface geometric details (Baseline (\textbf{\Romannum{1}})), whereas the RGB-D method provides overly smooth surface reconstruction (Baseline (\textbf{\Romannum{2}})). On the contrary, our method can reconstruct the fine surface details while maintaining the global shape structure.
Such results validate our idea of uncertainty guided policy for active 3D data acquisition using neural rendering principle.

\begin{figure}
\setlength{\abovecaptionskip}{-0.00cm}
\setlength{\belowcaptionskip}{-0.50cm}
    \centering
    \includegraphics[width=0.48\textwidth]{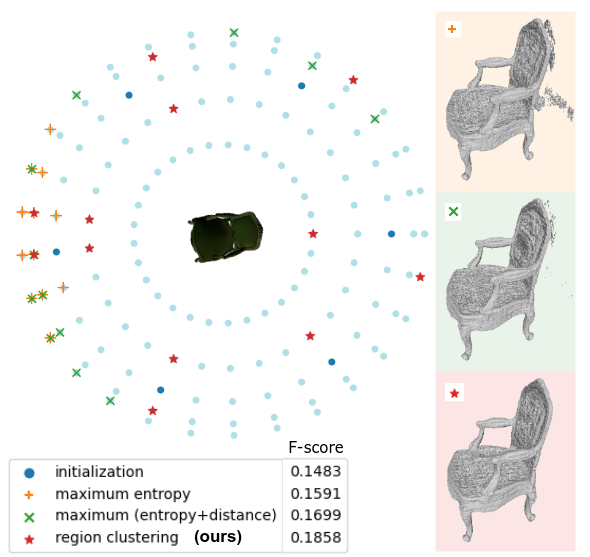}
    \caption{\textbf{Ablation study on view selection policies.} The left plot shows the selected poses using each policy, seen from above the hemisphere. Without the region clustering we proposed in Sec. \ref{subsec:policy}, selected poses have similar viewpoints with each other, resulting in less information gain and thus less precise reconstruction.}
    \label{fig:ablation_max}
\end{figure}

\subsection{Discussion on Runtime}
\noindent
The average initial training time of the NeRF \cite{mildenhall2020nerf} is around 15 hours when trained on a single NVIDIA GPU (TITAN Xp). Although current experiments cost a significant amount of time to train NeRF models, our uncertainty estimation is general and can be adopted in various NeRF-based approaches. To this end, we performed additional experiments using TensoRF \cite{tensorf}, a NeRF-based model that achieves fast training. The training time is about 8.2 minutes on a single NVIDIA 2080 GPU, and the F-scores from different policies when evaluated on \textit{Lego} are: Initialization (0.3272); Random (0.4895), Heuristic (0.4234), Similarity (0.4163), GT similarity (0.4598), Ours (0.5078); All Image (0.6281). Thus, our policy is model agnostic, and the proposed idea can be switched to an alternative NeRF-based model depending on the application. In addition, we report the computation time for the next-best-view selection using each policy: Random and Heuristic - less than a second, Similarity (GT) - 15 sec., VI \cite{scaramuzza} - 13.8 sec., Similarity and Ours - approximately 5 min. Further, a faster inference time can be achieved with TensoRF \cite{tensorf}, reducing the uncertainty computation time from 5 minutes to 2.8 minutes.

\vspace{-0.2cm}
\subsection{Ablation Study} \label{sec:ablation}
\noindent
\textbf{(a) View Selection Policy.}
As discussed in Sec. \ref{subsec:policy}, we use region clustering to avoid selecting the next views concentrated in a certain region. We show the proposed policy is an effective approach by comparing the reconstruction result to \textit{(i)} when we select the most uncertain views without considering other factors such as the locality (\textcolor{orange}{+} symbol in Fig.(\ref{fig:ablation_max})), and \textit{(ii)} when we consider the average spherical distances between each pair of chosen poses (\textcolor{Green}{$\times$} symbol in Fig.(\ref{fig:ablation_max})).
Fig.(\ref{fig:ablation_max}) shows the chosen next views for each policy and the resulting 3D meshes. If we use policy \textit{(i)}, all 12 selected views are next to each other and none of them contains the side or the rear view of the object. Therefore, the rear part of the reconstruction is particularly noisy. When policy \textit{(ii)} is used, the selected views are more distributed than in the former case but the rear part of the object is not yet precisely reconstructed. On  {the} contrary, when using our uncertainty guided policy (\textcolor{red}{$\filledstar$} symbol in Fig.(\ref{fig:ablation_max})), denoted as region clustering, we recover 3D mesh with better quality.

\smallskip
\noindent
\textbf{(b) Iterative Reconstruction.} When we run our active robotic 3D reconstruction pipeline, we show that we can achieve a comparable level of 3D reconstruction quality without using the whole image set. Fig.(\ref{fig:iterative}) shows the resulting 3D meshes of \emph{Chair} at different iterations. After one iteration, the noise floating behind the chair is filtered; however, the rear part of the backrest is still noisy, and the subtle convex shape of the seat and the front part of the backrest is not captured. After four iterations, these fine details are well represented, resulting in a mesh similar to the one we can get by using 150 images with only 54 images.  {In addition, we observed a decrease in the average of the mean entropy values over all pixels after each iteration: 1.748, 0.837, 0.797, 0.791, 0.790, which complies with our intuition.} Therefore, we can efficiently reconstruct an object by actively choosing the camera poses with our policy.

\newcommand{\resultscropheight}{0.75in}
\renewcommand{\resultscropwidth}{0.75in}

\newcommand{\cropchair}[2][1]{
  \makecell{
  \includegraphics[trim={460px 320px 500px 100px}, clip, height=\resultscropheight]{#2} \\
  \includegraphics[trim={420px 270px 350px 250px}, clip, width=\resultscropwidth]{#1}
  }
}

\begin{figure}[t]
\setlength{\abovecaptionskip}{0.225cm}
\setlength{\belowcaptionskip}{-0.5cm}
\centering
\scriptsize
\begin{tabular}{c@{}c@{}c@{}c@{}}
\cropchair[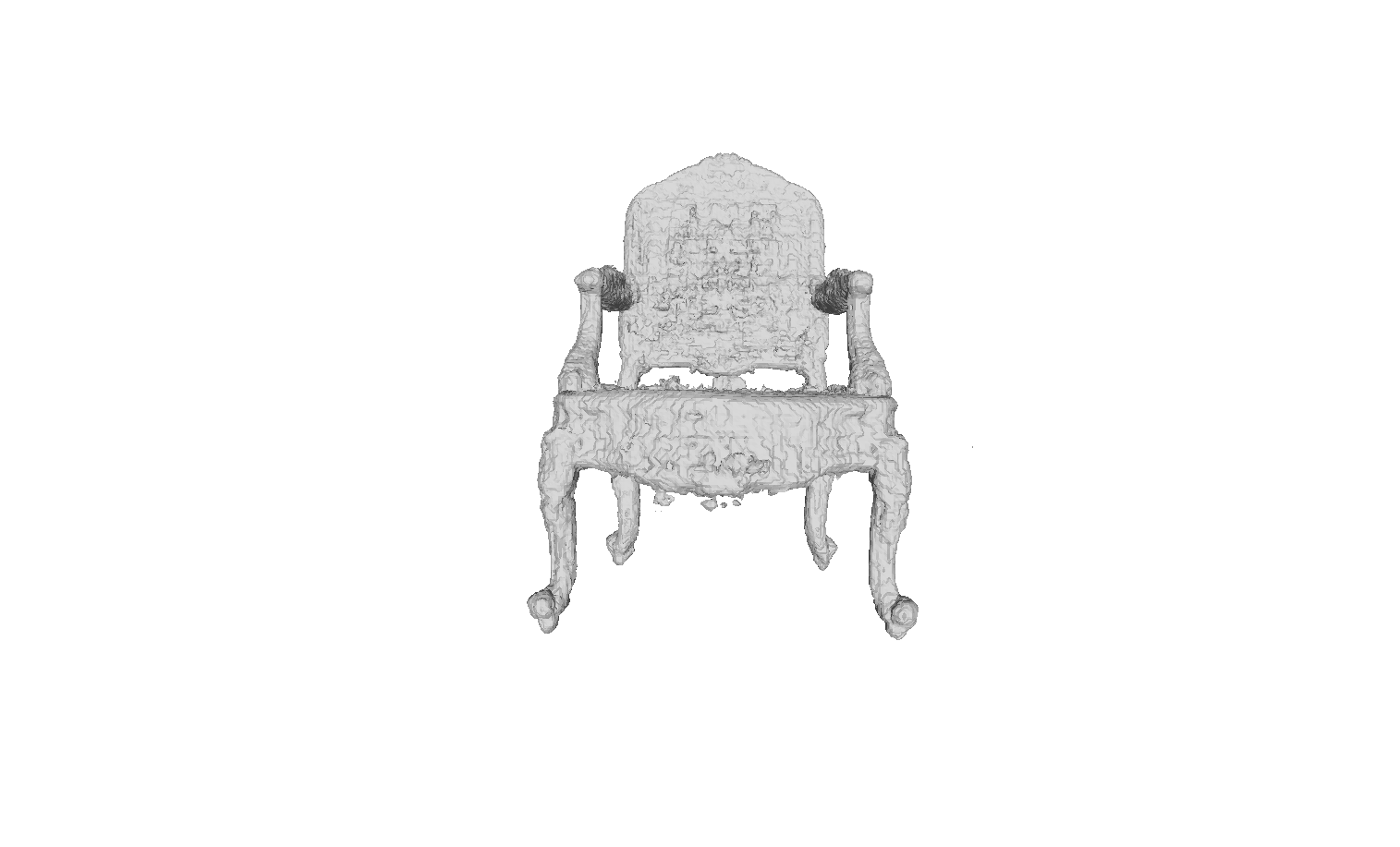]{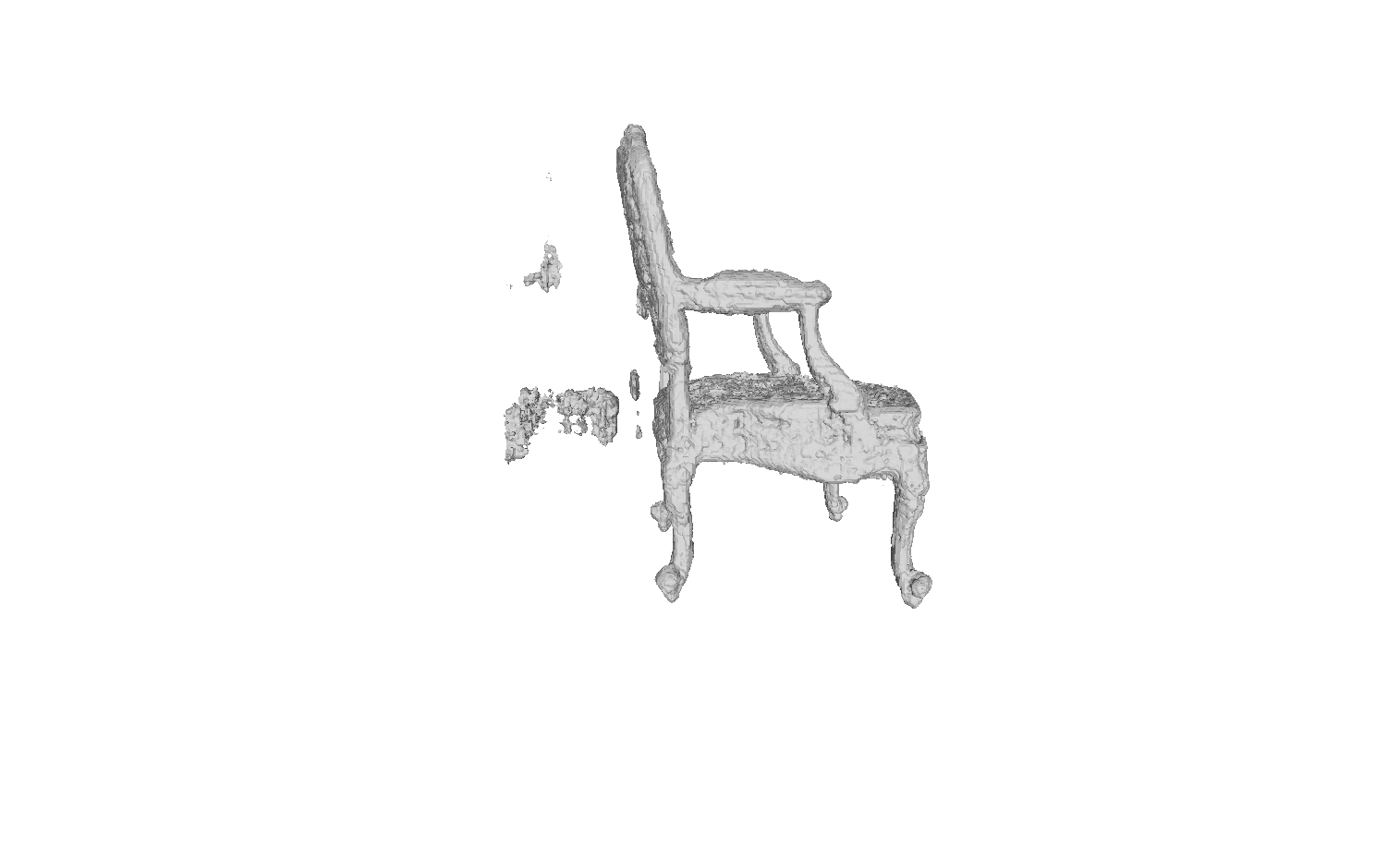} &
\cropchair[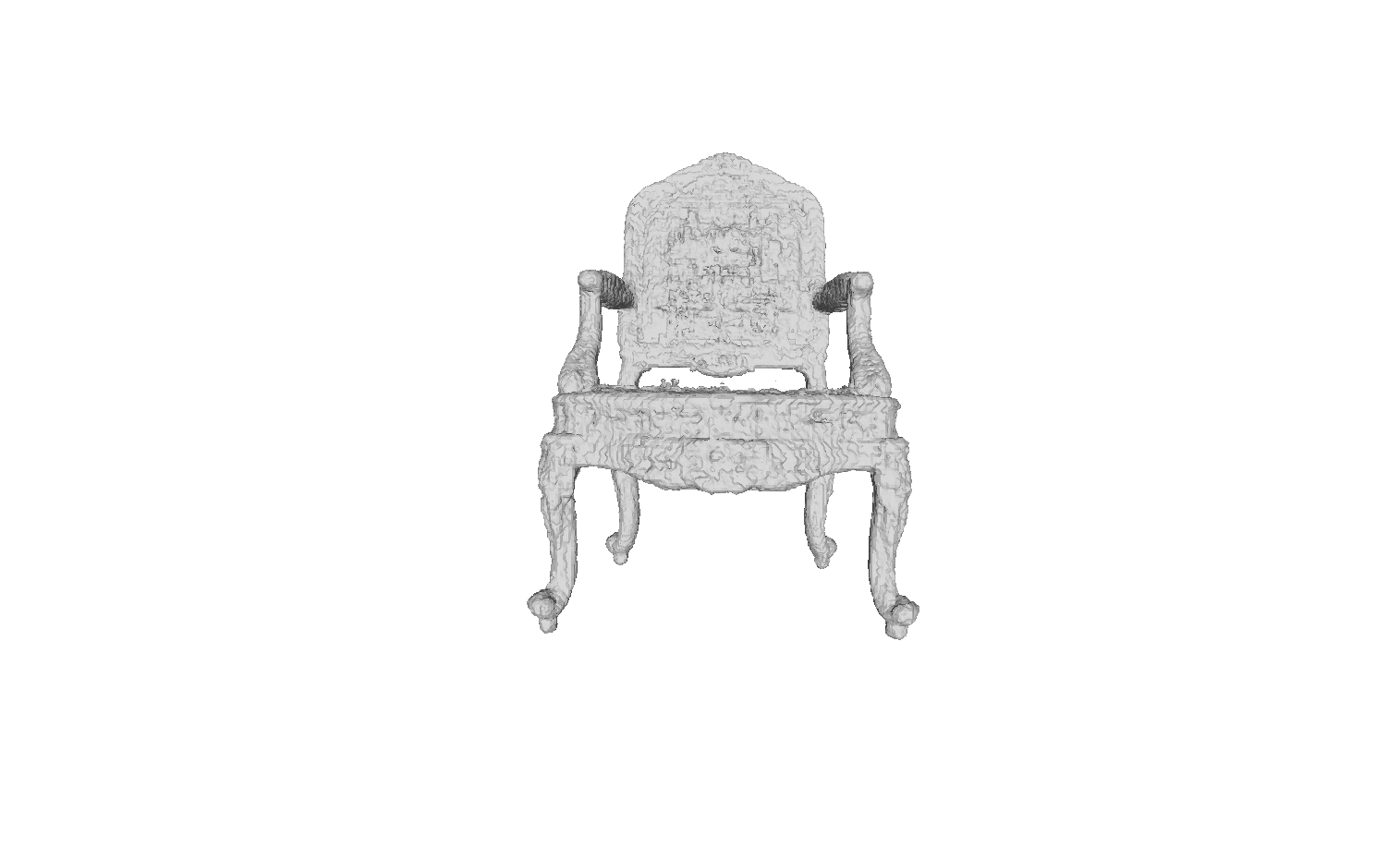]{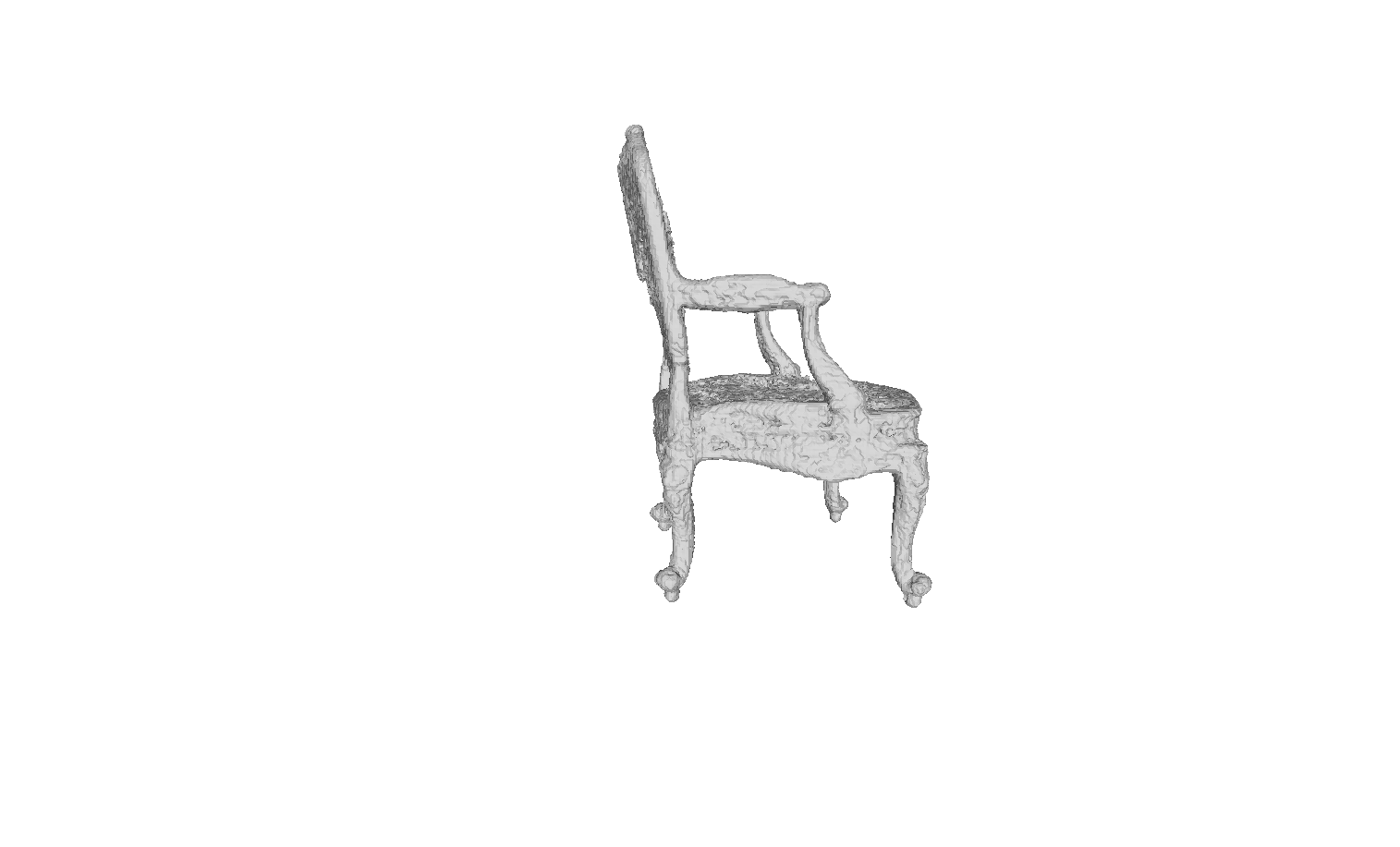} &
\cropchair[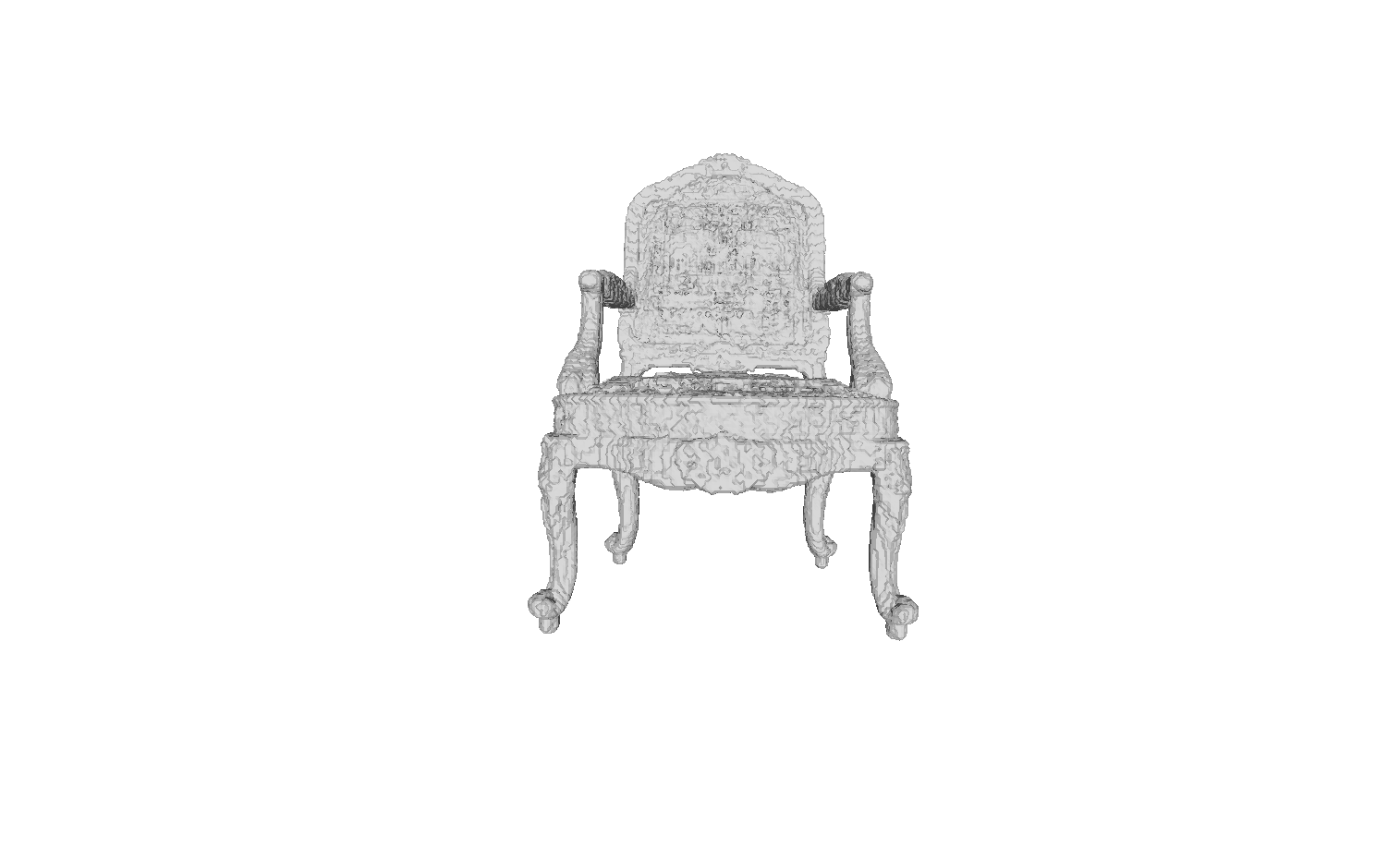]{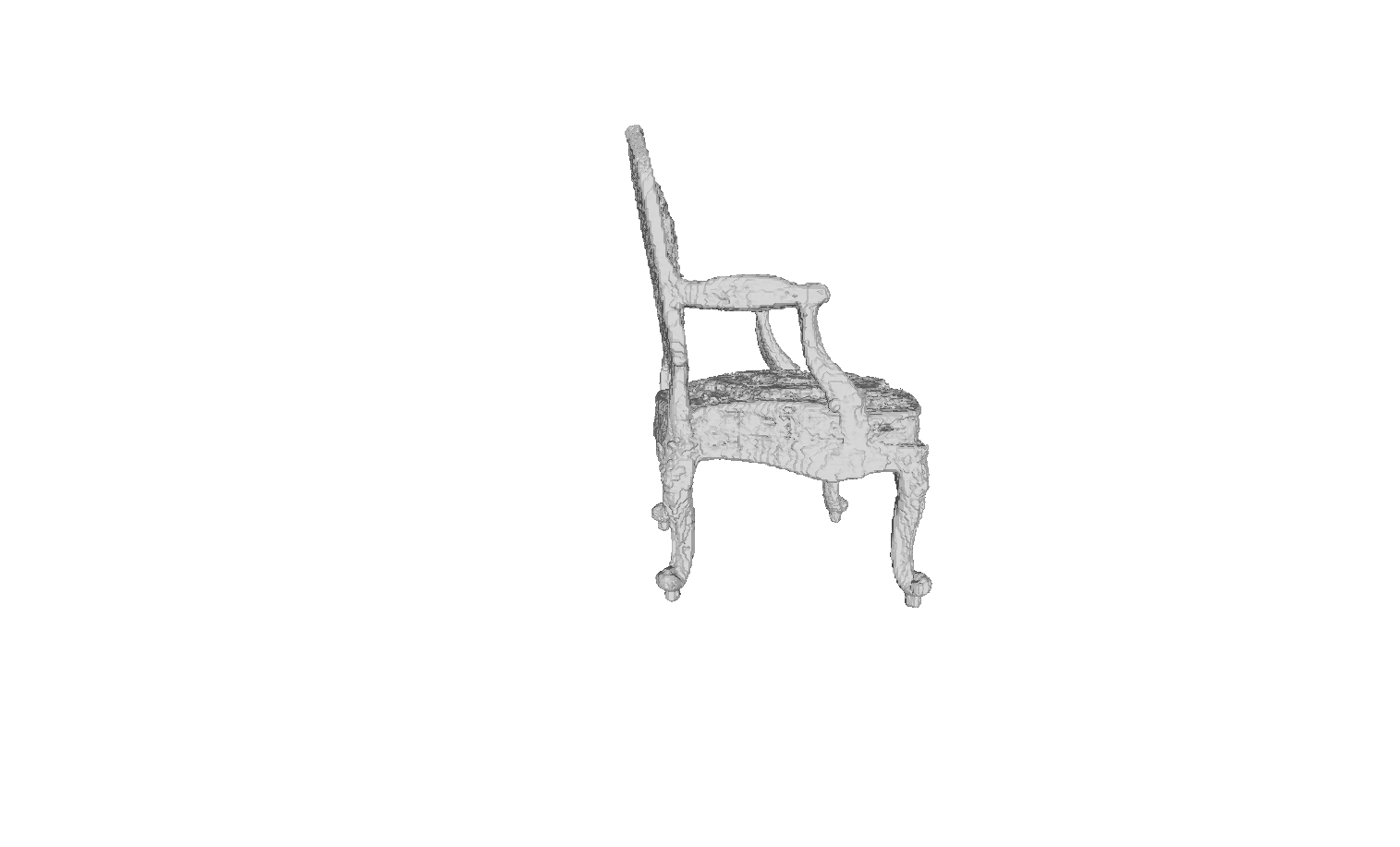} &
\cropchair[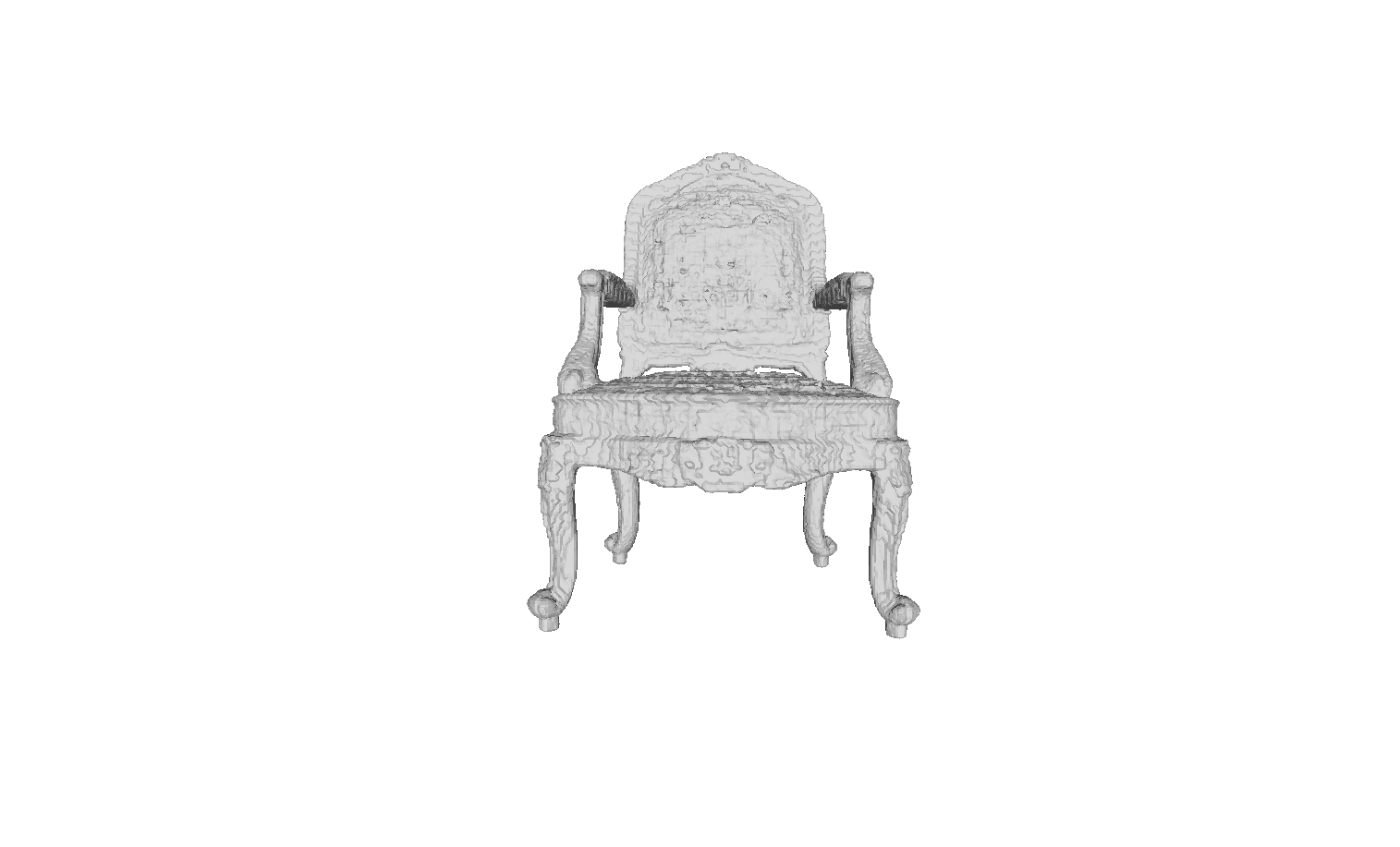]{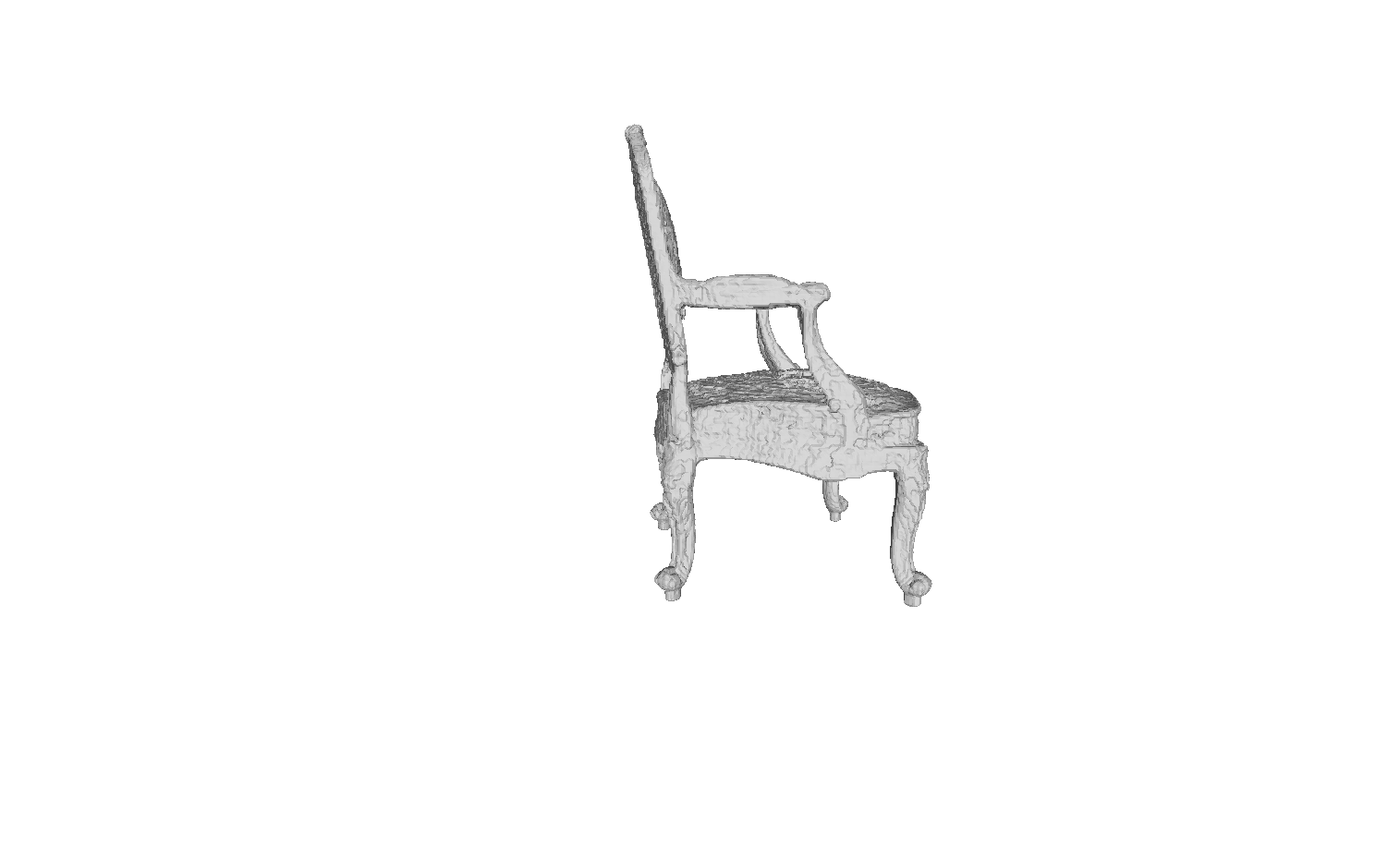}\\
Initialization & Iteration 1 & Iteration 4 & \textit{All Images}\\
(6 images) & (18 images) & (54 images) & (150 images)
\end{tabular}
\caption{\textbf{Iterative reconstruction result}. After 4 iterations using our uncertainty guided policy, the model is trained using only 54 images, but the mesh quality of the model is comparable to when all 150 images are used.}
\label{fig:iterative}
\end{figure}

\section{Conclusion}
\label{sec:conclusion}

This paper leveraged neural radiance fields-based implicit representations to tackle active robotic 3D reconstruction of an object. First, we introduced the ray-based volumetric uncertainty estimator, which provides a suitable proxy for the uncertainty of the underlying 3D geometry by computing the entropy of the weight distribution of color samples along rays. The proposed uncertainty estimation is applicable and can be generalized to other recently improved neural rendering-based approaches. Then, based on the estimator, we proposed an uncertainty-guided policy for the robotic system to determine the next best view for effective 3D reconstruction of an object. Experiments on synthetic and real-world examples show that our policy selects informative views for the object's better active 3D reconstruction.

Indeed, the optimization and rendering time of classical NeRF can be argued. Nevertheless, as tested using TensoRF \cite{tensorf} implementation, our method is general and can be further improved using advanced neural rendering methods with faster implementations. We believe our proposed method opens up a new research direction of using an implicit 3D object representation for the next-best-view selection problem in robot vision applications.

\bibliographystyle{IEEEtran} 
\bibliography{IEEEabrv.bib,reference.bib}

\begin{thebibliography}{10}
\providecommand{\url}[1]{#1}
\csname url@rmstyle\endcsname
\providecommand{\newblock}{\relax}
\providecommand{\bibinfo}[2]{#2}
\providecommand\BIBentrySTDinterwordspacing{\spaceskip=0pt\relax}
\providecommand\BIBentryALTinterwordstretchfactor{4}
\providecommand\BIBentryALTinterwordspacing{\spaceskip=\fontdimen2\font plus
\BIBentryALTinterwordstretchfactor\fontdimen3\font minus
  \fontdimen4\font\relax}
\providecommand\BIBforeignlanguage[2]{{%
\expandafter\ifx\csname l@#1\endcsname\relax
\typeout{** WARNING: IEEEtran.bst: No hyphenation pattern has been}%
\typeout{** loaded for the language `#1'. Using the pattern for}%
\typeout{** the default language instead.}%
\else
\language=\csname l@#1\endcsname
\fi
#2}}

\bibitem{peralta2020next}
D.~Peralta, J.~Casimiro, A.~M. Nilles, J.~A. Aguilar, R.~Atienza, and
  R.~Cajote, ``Next-best view policy for 3d reconstruction,'' in
  \emph{ECCV}.\hskip 1em plus 0.5em minus 0.4em\relax Springer, 2020, pp.
  558--573.

\bibitem{scaramuzza}
S.~Isler, R.~Sabzevari, J.~Delmerico, and D.~Scaramuzza, ``An information gain
  formulation for active volumetric 3d reconstruction,'' in \emph{ICRA}, 2016,
  pp. 3477--3484.

\bibitem{delmerico2018comparison}
J.~Delmerico, S.~Isler, R.~Sabzevari, and D.~Scaramuzza, ``A comparison of
  volumetric information gain metrics for active 3d object reconstruction,''
  \emph{Autonomous Robots}, vol.~42, no.~2, pp. 197--208, 2018.

\bibitem{Chen2011ActiveVI}
S.~Chen, Y.~Li, and N.~Kwok, ``Active vision in robotic systems: A survey of
  recent developments,'' \emph{The International Journal of Robotics Research},
  vol.~30, pp. 1343 -- 1377, 2011.

\bibitem{zeng2020view}
R.~Zeng, Y.~Wen, W.~Zhao, and Y.-J. Liu, ``View planning in robot active
  vision: A survey of systems, algorithms, and applications,''
  \emph{Computational Visual Media}, pp. 1--21, 2020.

\bibitem{devrim2017reinforcement}
M.~Devrim~Kaba, M.~Gokhan~Uzunbas, and S.~Nam~Lim, ``A reinforcement learning
  approach to the view planning problem,'' in \emph{CVPR}, 2017, pp.
  6933--6941.

\bibitem{mendoza2020supervised}
M.~Mendoza, J.~I. Vasquez-Gomez, H.~Taud, L.~E. Sucar, and C.~Reta,
  ``Supervised learning of the next-best-view for 3d object reconstruction,''
  \emph{Pattern Recognition Letters}, vol. 133, pp. 224--231, 2020.

\bibitem{wang2019autonomous}
Y.~Wang, S.~James, E.~K. Stathopoulou, C.~Beltr{\'a}n-Gonz{\'a}lez, Y.~Konishi,
  and A.~Del~Bue, ``Autonomous 3-d reconstruction, mapping, and exploration of
  indoor environments with a robotic arm,'' \emph{IEEE Robotics and Automation
  Letters}, vol.~4, no.~4, pp. 3340--3347, 2019.

\bibitem{wu2019plant}
C.~Wu, R.~Zeng, J.~Pan, C.~C. Wang, and Y.-J. Liu, ``Plant phenotyping by
  deep-learning-based planner for multi-robots,'' \emph{IEEE Robotics and
  Automation Letters}, vol.~4, no.~4, pp. 3113--3120, 2019.

\bibitem{wu2014poisson}
S.~Wu, W.~Sun, P.~Long, H.~Huang, D.~Cohen-Or, M.~Gong, O.~Deussen, and
  B.~Chen, ``Quality-driven poisson-guided autoscanning,'' \emph{ACM Trans.
  Graph.}, vol.~33, no.~6, Nov. 2014.

\bibitem{mildenhall2020nerf}
B.~Mildenhall, P.~P. Srinivasan, M.~Tancik, J.~T. Barron, R.~Ramamoorthi, and
  R.~Ng, ``{NeRF}: Representing scenes as neural radiance fields for view
  synthesis,'' in \emph{ECCV}, 2020.

\bibitem{bianco2018evaluating}
S.~Bianco, G.~Ciocca, and D.~Marelli, ``Evaluating the performance of structure
  from motion pipelines,'' \emph{Journal of Imaging}, vol.~4, no.~8, p.~98,
  2018.

\bibitem{schoenberger2016sfm}
J.~L. Sch\"{o}nberger and J.-M. Frahm, ``Structure-from-motion revisited,'' in
  \emph{CVPR}, 2016.

\bibitem{furukawa2009accurate}
Y.~Furukawa and J.~Ponce, ``Accurate, dense, and robust multiview stereopsis,''
  \emph{IEEE T-PAMI}, vol.~32, no.~8, pp. 1362--1376, 2009.

\bibitem{curless1996volumetric}
B.~Curless and M.~Levoy, ``A volumetric method for building complex models from
  range images,'' in \emph{Proceedings of the 23rd annual conference on
  Computer graphics and interactive techniques}, 1996, pp. 303--312.

\bibitem{choi2015robust}
S.~Choi, Q.-Y. Zhou, and V.~Koltun, ``Robust reconstruction of indoor scenes,''
  in \emph{CVPR}, 2015, pp. 5556--5565.

\bibitem{newcombe2011kinectfusion}
R.~A. Newcombe, S.~Izadi, O.~Hilliges, D.~Molyneaux, D.~Kim, A.~J. Davison,
  P.~Kohi, J.~Shotton, S.~Hodges, and A.~Fitzgibbon, ``Kinectfusion: Real-time
  dense surface mapping and tracking,'' in \emph{2011 10th IEEE international
  symposium on mixed and augmented reality}.\hskip 1em plus 0.5em minus
  0.4em\relax IEEE, 2011, pp. 127--136.

\bibitem{wu2011visualsfm}
C.~Wu \emph{et~al.}, ``Visualsfm: A visual structure from motion system,''
  2011.

\bibitem{yuan2018pcn}
W.~Yuan, T.~Khot, D.~Held, C.~Mertz, and M.~Hebert, ``Pcn: Point completion
  network,'' in \emph{3DV}, 2018.

\bibitem{icml2020_2086}
A.~Gropp, L.~Yariv, N.~Haim, M.~Atzmon, and Y.~Lipman, ``Implicit geometric
  regularization for learning shapes,'' in \emph{Proceedings of Machine
  Learning and Systems 2020}, 2020, pp. 3569--3579.

\bibitem{Takikawa2021nglod}
T.~Takikawa, J.~Litalien, K.~Yin, K.~Kreis, C.~Loop, D.~Nowrouzezahrai,
  A.~Jacobson, M.~McGuire, and S.~Fidler, ``Neural geometric level of detail:
  Real-time rendering with implicit 3d shapes,'' in \emph{CVPR}, 2021, pp.
  11\,358--11\,367.

\bibitem{mescheder2019occupancy}
L.~Mescheder, M.~Oechsle, M.~Niemeyer, S.~Nowozin, and A.~Geiger, ``Occupancy
  networks: Learning 3d reconstruction in function space,'' in \emph{CVPR},
  2019, pp. 4460--4470.

\bibitem{park2019deepsdf}
J.~J. Park, P.~Florence, J.~Straub, R.~Newcombe, and S.~Lovegrove, ``Deepsdf:
  Learning continuous signed distance functions for shape representation,'' in
  \emph{CVPR}, 2019, pp. 165--174.

\bibitem{SucarICCV2021imap}
E.~Sucar, S.~Liu, J.~Ortiz, and A.~Davison, ``{iMAP}: Implicit mapping and
  positioning in real-time,'' in \emph{ICCV}, 2021.

\bibitem{schoenberger2016mvs}
J.~L. Sch\"{o}nberger, E.~Zheng, M.~Pollefeys, and J.-M. Frahm, ``Pixelwise
  view selection for unstructured multi-view stereo,'' in \emph{ECCV}, 2016.

\bibitem{niemeyer2020differentiable}
M.~Niemeyer, L.~Mescheder, M.~Oechsle, and A.~Geiger, ``Differentiable
  volumetric rendering: Learning implicit 3d representations without 3d
  supervision,'' in \emph{CVPR}, 2020, pp. 3504--3515.

\bibitem{yariv2020multiview}
L.~Yariv, Y.~Kasten, D.~Moran, M.~Galun, M.~Atzmon, B.~Ronen, and Y.~Lipman,
  ``Multiview neural surface reconstruction by disentangling geometry and
  appearance,'' \emph{NeurIPS}, vol.~33, 2020.

\bibitem{oechsle2021unisurf}
M.~Oechsle, S.~Peng, and A.~Geiger, ``Unisurf: Unifying neural implicit
  surfaces and radiance fields for multi-view reconstruction,'' in \emph{ICCV},
  2021, pp. 5589--5599.

\bibitem{wang2021neus}
P.~Wang, L.~Liu, Y.~Liu, C.~Theobalt, T.~Komura, and W.~Wang, ``Neus: Learning
  neural implicit surfaces by volume rendering for multi-view reconstruction,''
  \emph{NeurIPS}, 2021.

\bibitem{yariv2021volume}
L.~Yariv, J.~Gu, Y.~Kasten, and Y.~Lipman, ``Volume rendering of neural
  implicit surfaces,'' \emph{NeurIPS}, vol.~34, 2021.

\bibitem{yu2021pixelnerf}
A.~Yu, V.~Ye, M.~Tancik, and A.~Kanazawa, ``pixelnerf: Neural radiance fields
  from one or few images,'' in \emph{CVPR}, 2021, pp. 4578--4587.

\bibitem{tensorf}
A.~Chen, Z.~Xu, A.~Geiger, J.~Yu, and H.~Su, ``Tensorf: Tensorial radiance
  fields,'' \emph{arXiv preprint arXiv:2203.09517}, 2022.

\bibitem{wang2021nerfmm}
Z.~Wang, S.~Wu, W.~Xie, M.~Chen, and V.~A. Prisacariu, ``Ne{RF}$--$: Neural
  radiance fields without known camera parameters,'' \emph{arXiv preprint
  arXiv:2102.07064}, 2021.

\bibitem{lin2021barf}
C.-H. Lin, W.-C. Ma, A.~Torralba, and S.~Lucey, ``Barf: Bundle-adjusting neural
  radiance fields,'' in \emph{ICCV}, 2021.

\bibitem{kangle2021dsnerf}
K.~Deng, A.~Liu, J.-Y. Zhu, and D.~Ramanan, ``Depth-supervised {NeRF}: Fewer
  views and faster training for free,'' in \emph{CVPR}, June 2022.

\bibitem{martin2021nerfw}
R.~Martin-Brualla, N.~Radwan, M.~S. Sajjadi, J.~T. Barron, A.~Dosovitskiy, and
  D.~Duckworth, ``Nerf in the wild: Neural radiance fields for unconstrained
  photo collections,'' in \emph{CVPR}, 2021, pp. 7210--7219.

\bibitem{shen2021stochastic}
J.~Shen, A.~Ruiz, A.~Agudo, and F.~Moreno-Noguer, ``Stochastic neural radiance
  fields: Quantifying uncertainty in implicit 3d representations,'' in
  \emph{3DV}.\hskip 1em plus 0.5em minus 0.4em\relax IEEE, 2021, pp. 972--981.

\bibitem{kajiya1984ray}
J.~T. Kajiya and B.~P. Von~Herzen, ``Ray tracing volume densities,'' \emph{ACM
  SIGGRAPH computer graphics}, vol.~18, no.~3, pp. 165--174, 1984.

\bibitem{max1995}
N.~Max, ``Optical models for direct volume rendering,'' \emph{IEEE Transactions
  on Visualization and Computer Graphics}, vol.~1, no.~2, pp. 99--108, 1995.

\bibitem{mcubes}
W.~E. Lorensen and H.~E. Cline, ``Marching cubes: A high resolution 3d surface
  construction algorithm,'' in \emph{Proceedings of the 14th Annual Conference
  on Computer Graphics and Interactive Techniques}, ser. SIGGRAPH '87.\hskip
  1em plus 0.5em minus 0.4em\relax New York, NY, USA: Association for Computing
  Machinery, 1987, p. 163–169.

\bibitem{Knapitsch2017tank}
A.~Knapitsch, J.~Park, Q.-Y. Zhou, and V.~Koltun, ``Tanks and temples:
  Benchmarking large-scale scene reconstruction,'' \emph{ACM Transactions on
  Graphics}, vol.~36, no.~4, 2017.

\bibitem{gupta2018robot}
A.~Gupta, A.~Murali, D.~P. Gandhi, and L.~Pinto, ``Robot learning in homes:
  Improving generalization and reducing dataset bias,'' \emph{NeurIPS},
  vol.~31, 2018.

\bibitem{apriltag}
E.~Olson, ``{AprilTag}: A robust and flexible visual fiducial system,'' in
  \emph{2011 IEEE ICRA}, 2011, pp. 3400--3407.

\bibitem{he2016deep}
K.~He, X.~Zhang, S.~Ren, and J.~Sun, ``Deep residual learning for image
  recognition,'' in \emph{CVPR}, 2016, pp. 770--778.

\bibitem{deng2009imagenet}
J.~Deng, W.~Dong, R.~Socher, L.-J. Li, K.~Li, and L.~Fei-Fei, ``Imagenet: A
  large-scale hierarchical image database,'' in \emph{CVPR}, 2009, pp.
  248--255.

\bibitem{poisson}
M.~Kazhdan and H.~Hoppe, ``Screened poisson surface reconstruction,'' \emph{ACM
  Trans. Graph.}, vol.~32, no.~3, July 2013.

\bibitem{Park2017ColoredPC}
J.~Park, Q.-Y. Zhou, and V.~Koltun, ``Colored point cloud registration
  revisited,'' \emph{ICCV}, pp. 143--152, 2017.

\end{thebibliography}

\end{document}


\maketitle
\appendices

\thispagestyle{empty}
\pagestyle{empty}



\section{Robotic System Design}
We use LoCoBot, a low-cost mobile manipulator platform~\cite{gupta2018robot}, as our robotic system for active 3D reconstruction. As mentioned in the main paper, the robot consists of a wheel-based mobile base that has two degrees of freedom (DoF), a manipulator with 5 DoF, and an RGB-D camera (Intel RealSense D435) attached to the top. An Intel NUC Kit NUC8i3BEH serves as the main processor of the system. The original LoCoBot has a gripper at the end of the manipulator. However, in our robot setup, the gripper is replaced with another RGB-D camera (Intel RealSense D415) to give the robot the capability to scan details for better 3D reconstructions. The adjusted hardware is shown in Fig~\ref{fig:locobot}.

\begin{figure}[htb]
    \centering
    \setlength{\belowcaptionskip}{-0.1cm}
    \includegraphics[width=0.38\textwidth]{images/007_locobot_compressed_with_label.png}
    \caption[Hardware platform]{The hardware platform used for our active robotic 3D reconstruction.}
    \label{fig:locobot}
\end{figure}

We utilize APIs built with the Robot Operating System (ROS)~\cite{ros}  provided by Trossen Robotics\footnote{BSD 2-Clause License.}. Some of the high-level functionalities that are implemented on top of the base APIs originate from PyRobot\footnote{MIT License.}~\cite{pyrobot2019}, which are modified and adjusted to fit our needs.

\section{Additional Visualization of Results}


We show additional qualitative results in Fig.~\ref{fig:supp_partial_meshes}.
The first row of the result on \textit{Lego} shows the reconstruction of the upper back of the model. The red rectangles highlight the downward plane of the arch. The one reconstructed by our method has smooth planes and straight lines, being similar to the one reconstructed with all images. However, for the results of other policies, there are many protuberances on the surfaces. In addition, the second row of \textit{Lego} shows the reconstruction results of the caterpillar tracks of the loader. Compared with other policies, the caterpillar track reconstructed by our method is much more complete and has more details.

We also present the result on \textit{Drums}. The first row shows the beater of the bass drum pedal. While it is fully reconstructed when all 150 images are used, it is often ignored when we train a model with 6 or 18 images. Still, the shape is retained the best when using our method compared to the other baseline methods. The second row shows the interior surface of the high tom. The surface is noisy and looks undesirably thick when reconstructed using the baseline methods. On the contrary, the surface is closer to an ideal surface when using our method.
This stands in contrast to the baseline methods, which achieve a reasonable overall reconstruction but perform worse in regions with fine details and parts that need well-placed views to reduce the effect of occlusions. 

\newcommand{\resultsfigwidth}{0.9in}
\newcommand{\resultscropwidth}{0.69in}

\newcommand{\croplego}[2][1]{
  \makecell{
  \includegraphics[trim={220px 140px 130px 40px}, clip, width=\resultscropwidth]{#2} \\
  \includegraphics[trim={180px 120px 200px 90px}, clip, width=\resultscropwidth]{#1}
  }
}

\newcommand{\cropdrum}[2][1]{
  \makecell{
  \includegraphics[trim={220px 140px 130px 40px}, clip, width=\resultscropwidth]{#2} \\
  \includegraphics[trim={180px 120px 200px 90px}, clip, width=\resultscropwidth]{#1}
  }
}

\begin{figure*}[t]
\setlength{\belowcaptionskip}{0cm}
\centering
\scriptsize
\begin{tabular}{@{}c@{}c@{}c@{}c@{}c@{}c@{}c@{}c@{}c@{}}
\makecell[c]{
\includegraphics[trim={0px 50px 0px 50px}, clip, width=\resultsfigwidth]{images/009_lego.png}
\\
Lego
}
&
\croplego[images/lego_compressed/init.png]{images/lego_compressed/init_2.png} &
\croplego[images/lego_compressed/random.png]{images/lego_compressed/random_2.png} &
\croplego[images/lego_compressed/human.png]{images/lego_compressed/human_2.png} &
\croplego[images/lego_compressed/image_feature_render.png]{images/lego_compressed/image_feature_render_2.png} &
\croplego[images/lego_compressed/image_feature.png]{images/lego_compressed/image_feature_2.png} &
\croplego[images/lego_compressed/rpg.png]{images/lego_compressed/rpg_2.png} &
\croplego[images/lego_compressed/entropy.png]{images/lego_compressed/entropy_2.png} &
\croplego[images/lego_compressed/baseline.png]{images/lego_compressed/baseline_2.png} \\

\makecell[c]{
\includegraphics[trim={0px 50px 0px 50px}, clip, width=\resultsfigwidth]{images/009_drum.png}
\\
Drums
}
&
\cropdrum[images/drum_compressed/init.png]{images/drum_compressed/2_init.png} &
\cropdrum[images/drum_compressed/random.png]{images/drum_compressed/2_random.png} &
\cropdrum[images/drum_compressed/heuristic.png]{images/drum_compressed/2_heuristic.png} &
\cropdrum[images/drum_compressed/sim.png]{images/drum_compressed/2_sim.png} &
\cropdrum[images/drum_compressed/sim_gt.png]{images/drum_compressed/2_sim_gt.png} &
\cropdrum[images/drum_compressed/vi.png]{images/drum_compressed/2_vi.png} &
\cropdrum[images/drum_compressed/ours.png]{images/drum_compressed/2_ours.png} &
\cropdrum[images/drum_compressed/all.png]{images/drum_compressed/2_all.png} \\

& Initialization & Random & Heuristic & Similarity & Similarity (GT) & VI~\cite{scaramuzza} & Ours & \textit{All Images}
\end{tabular}
\caption{\textbf{Comparisons on 3D meshes for \textit{Lego} and \textit{Drums}.} We illustrate a qualitative comparison against the baselines for the reconstructed geometry on the other objects that have high frequency details. Our method captures the fine geometry well compared to the other policies. Note that the VI \cite{scaramuzza} method yields voxel representations.}
\label{fig:supp_partial_meshes}
\end{figure*}

\renewcommand{\resultscropwidth}{1.56in}

\begin{figure*}
\setlength{\belowcaptionskip}{-0.3cm}
\centering
\scriptsize
\begin{tabular}{c@{}c@{}c@{}c@{}c@{}}
Initial &
\raisebox{-.5\height}{
\includegraphics[trim={180px 100px 180px 50px}, clip, width=\resultscropwidth]{images/full_mesh/lego_init00.png}} &
\raisebox{-.5\height}{
\includegraphics[trim={150px 100px 140px 20px}, clip, width=\resultscropwidth]{images/full_mesh/chair_init00.png}} &
\raisebox{-.5\height}{
\includegraphics[trim={240px 100px 280px 100px}, clip, width=\resultscropwidth]{images/full_mesh/supp_drum_init01.png}} &
\raisebox{-.5\height}{
\includegraphics[trim={300px 120px 280px 80px}, clip, width=\resultscropwidth]{images/full_mesh/supp_ficus_init02.png}} \\
Ours &
\raisebox{-.5\height}{
\includegraphics[trim={180px 100px 180px 50px}, clip, width=\resultscropwidth]{images/full_mesh/lego_entropy00.png}} &
\raisebox{-.5\height}{
 \includegraphics[trim={150px 100px 140px 20px}, clip, width=\resultscropwidth]{images/full_mesh/chair_entropy00.png}} &
\raisebox{-.5\height}{
\includegraphics[trim={240px 100px 280px 100px}, clip, width=\resultscropwidth]{images/full_mesh/supp_drum_ours00.png}} &
\raisebox{-.5\height}{
\includegraphics[trim={300px 120px 280px 80px}, clip, width=\resultscropwidth]{images/full_mesh/supp_ficus_ours01.png}} \\
Ground Truth &
\raisebox{-.5\height}{
\includegraphics[trim={180px 100px 180px 50px}, clip, width=\resultscropwidth]{images/full_mesh/lego_gt00.png}} &
\raisebox{-.5\height}{
\includegraphics[trim={150px 100px 140px 20px}, clip, width=\resultscropwidth]{images/full_mesh/chair_gt00.png}} &
\raisebox{-.5\height}{
\includegraphics[trim={240px 100px 280px 100px}, clip, width=\resultscropwidth]{images/full_mesh/supp_drum_gt02.png}} &
\raisebox{-.5\height}{
\includegraphics[trim={300px 120px 280px 80px}, clip, width=\resultscropwidth]{images/full_mesh/supp_ficus_gt00.png}} \\
& \textit{Lego} & \textit{Chair} & \textit{Drums} & \textit{Ficus}
\end{tabular}
\caption{\textbf{Visualization of 3D Meshes.} We show the 3D mesh of \textit{Lego}, \textit{Chair}, \textit{Drums}, and \textit{Ficus} obtained using our method, together with the initial 3D mesh and the ground truth 3D mesh. Our uncertainty guided active robotic 3D reconstruction approach yields high-quality reconstructions of complex objects.}
\label{fig:supp_full_meshes}
\end{figure*}

Lastly, we show the global 
%
%
3D mesh of our four test cases in Fig.~\ref{fig:supp_full_meshes}. The first row shows the 3D meshes obtained from the initial models, that are trained using 6 images. The second row shows the 3D meshes from models that are trained using 18 images, which have 12 additional images selected by our proposed method. The last row shows the ground truth 3D meshes for comparisons. We demonstrate that our uncertainty guided active robotic 3D reconstruction approach is able to reconstruct the fine geometry of objects.

\section{Evaluation Protocol}
In our synthetic experiments, we compute the F-score for all the objects using a threshold of 0.003. Furthermore, we always use a grid of dimension $256\times256\times256$, with a side length of 2.4 when extracting a 3D mesh using the Marching Cube algorithm \cite{mcubes}. However, the different examples use different threshold parameters to deal with the difference in the reconstructed objects. Specifically, we use the following parameters: 60, 20, 60, 10 for \textit{Lego}, \textit{Chair}, \textit{Drums}, and \textit{Ficus}, respectively. In the real-world experiment, we use 100 as the threshold parameter for the Marching Cube algorithm \cite{mcubes}, and use a grid of dimension $256\times256\times256$ with a side length of 0.6.


\bibliographystyle{IEEEtran} 
\bibliography{IEEEabrv.bib,reference.bib}